\documentclass[pdflatex, sn-nature, oneside]{sn-jnl}

\makeatletter
\def\headerps@out{}
\makeatother

\usepackage{graphicx}%
\usepackage{multirow}%
\usepackage{amsmath,amssymb,amsfonts}%
\usepackage{amsthm}%
\usepackage{mathrsfs}%
\usepackage[title]{appendix}%
\usepackage[table]{xcolor}%
\usepackage{textcomp}%
\usepackage{manyfoot}%
\usepackage{tabularx}
\usepackage{caption}
\usepackage{enumitem}
\usepackage{booktabs}%
\usepackage{algorithm}%
\usepackage{algorithmicx}%
\usepackage{algpseudocode}%
\usepackage{listings}%
\usepackage[most]{tcolorbox}
\usepackage[skip=0.5\baselineskip]{parskip}
\usepackage{makecell}
\usepackage{newunicodechar}
\usepackage{wasysym}
\usepackage[normalem]{ulem}
\usepackage{relsize}
\usepackage{lineno}

\theoremstyle{thmstyleone}%
\theoremstyle{thmstyletwo}%

\theoremstyle{thmstylethree}%

\newtcolorbox[auto counter]{promptbox}[2][]{
    sharp corners,
    fonttitle=\bfseries\small,
    title={Prompt \thetcbcounter: #2}, 
    colback=blue!2!white,         
    colframe=blue!40!gray!60!black, 
    breakable, 
    before upper=\vspace{-0.5em}, 
    #1 
}

\lstdefinestyle{promptcontent}{
    basicstyle=\ttfamily\small,
    breaklines=true,
    breakindent=0pt,        
    breakatwhitespace=false, 
    columns=fullflexible,
    keepspaces=true,
}

\DeclareCaptionLabelSeparator{vline}{ $|$ } 

\newcommand{\LR}{\operatorname{LR}}
\newcommand{\IC}{\operatorname{IC}}
\newcommand{\MICA}{\operatorname{MICA}}
\newcommand{\Sim}{\operatorname{Sim}}
\newcommand{\Odds}{\operatorname{Odds}}
\newcommand{\Anno}{\operatorname{Anno}}
\newcommand{\Desc}{\operatorname{Desc}}
\newcommand{\Anc}{\operatorname{Anc}}

\newunicodechar{●}{\ensuremath{\CIRCLE}}
\newunicodechar{○}{\ensuremath{\Circle}}
\newunicodechar{◐}{\ensuremath{\LEFTcircle}}

\definecolor{tablegray}{gray}{0.92}
\definecolor{yellowheader}{HTML}{FFE699} 
\definecolor{yellowrow}{HTML}{FFF2CC}    
\definecolor{blueheader}{HTML}{BDD7EE}   
\definecolor{bluerow}{HTML}{DDEBF7}      

\renewcommand{\abstractname}{}

\setcitestyle{super,open={},close={}}

\begin{document}
\title[FusionRare]{\vspace{-3em}\textbf{Unified Multiparadigm Fusion Intelligence for Rare Disease Diagnosis}\vspace{-0.5em}}

\author{
    \relsize{-0.5}
    \fnm{Jian} \sur{Yang}\textsuperscript{1,\dag,*}, 
    \fnm{Peilin} \sur{Liu}\textsuperscript{2,\dag}, 
    \fnm{Songda} \sur{Li}\textsuperscript{2}, 
    \fnm{Yu} \sur{Cao}\textsuperscript{1},
    \fnm{Yuqing} \sur{Feng}\textsuperscript{1},
    \fnm{Lan} \sur{Yu}\textsuperscript{1},
    \fnm{Jiarong} \sur{Pan}\textsuperscript{1},
    \fnm{Jing} \sur{Ye}\textsuperscript{1},
    \fnm{Xinwen} \sur{Huang}\textsuperscript{1},
    \fnm{Chaochun} \sur{Zou}\textsuperscript{1},
    \fnm{Xudong} \sur{Lu}\textsuperscript{2},
    \fnm{Huilong} \sur{Duan}\textsuperscript{2},
    \fnm{and Haomin} \sur{Li}\textsuperscript{1,2,*}
    \vspace{-0.5em}
}

\affil[1]{\small \orgdiv{The Children's Hospital}, \orgname{Zhejiang University School of Medicine}, \orgaddress{\city{Hangzhou}, \postcode{310052}, \country{China}}\vspace{-0.3em}}

\affil[2]{\small \orgdiv{The College of Biomedical Engineering and Instrument Science}, \orgname{Zhejiang University}, \orgaddress{\city{Hangzhou}, \postcode{310027}, \country{China}}\vspace{-0.3em}}

\affil[]{\small \textsuperscript{\dag}These authors contributed equally to this work\vspace{-0.3em}}

\affil[]{\small \textsuperscript{*}Corresponding Author: hmli@zju.edu.cn (H. L.); yangjian\_bme@zju.edu.cn (J. Y.)\vspace{-5em}}

\abstract{Rare diseases (RDs) affect an estimated 300 million people globally\cite{the_lancet_global_health_landscape_2024}, leaving patients to navigate a protracted ``diagnostic odyssey'' constrained by clinical scarcity and fragmented medical knowledge\cite{faye_time_2024}. Artificial intelligence (AI) has progressively advanced RD diagnostics, from ontology-driven symbolic methods and deep-learning classifiers to large language model (LLM)-based systems\cite{zhao_agentic_2026}. However, each paradigm carries structural limitations---symbolic methods are bounded by knowledge completeness, neural models sacrifice interpretability, and LLM-based systems are vulnerable to systematic evaluation biases\cite{ghassemi_false_2021,zech_variable_2018,cheng_survey_2025}, particularly data leakage and contamination\cite{cheng_survey_2025}.
In this study, we introduce FusionRare, a unified multiparadigm fusion framework integrating knowledge-driven symbolic reasoning, deep-learning pattern recognition, case-based patient retrieval, and LLM-based heuristic inference through statistical meta-aggregation and knowledge-enhanced agentic reflection. Evaluated on 2,528 strictly isolated clinical cases, FusionRare achieves state-of-the-art RD prioritization, attaining a macro-averaged Recall@1 of 51\% for phenotype-only diagnosis and 79\% for phenotype–genotype joint analysis. These results represent decisive performance advantages over the advanced agentic system DeepRare\cite{zhao_agentic_2026}, the frontier LLMs including GPT-5.5\cite{openai_introducing_2026}, and specialized clinical tools such as PhenoBrain\cite{mao_phenotype-based_2025} and LIRICAL\cite{robinson_interpretable_2020} under stringent anti-leakage protocols.
A randomized, crossover clinician-AI validation further demonstrated that FusionRare elevates the true diagnostic rank in 45\% of decisions, improves physician confidence in 62\% of decisions, and reduces time to diagnosis by 21\%. These findings indicate that reconciling these orthogonal AI paradigms can significantly enhance comprehensive diagnostic capabilities, which is the key to bridging the gap between computational benchmarks and actionable clinical reliability. More broadly, this work provides a replicable framework for deploying trustworthy diagnostic AI across complex, high-stakes, and underserved medical domains.}

\maketitle

\section{Introduction}\label{sec1}
Rare diseases (RDs) collectively affect an estimated 300 million people worldwide; however, the extreme scarcity of individual conditions, which are typically defined by a prevalence of fewer than 1 in 2,000 individuals, often relegates them to the periphery of conventional medical training and healthcare infrastructure\cite{the_lancet_global_health_landscape_2024}. Although more than 7,000 distinct disorders have been documented, 72--80\% of which have a genetic aetiology, the trajectory towards clinical certainty remains fraught with systemic barriers\cite{the_lancet_global_health_landscape_2024,wakap_estimating_2020}. Patients typically endure a ``diagnostic odyssey'' spanning an average of five years, characterized by a cycle of redundant investigations and repeated misdiagnoses\cite{faye_time_2024}. This prolonged delay frequently leads to irreversible disease progression and misses critical therapeutic windows, resulting in profound psychological and financial burdens on families\cite{bauskis_diagnostic_2022}. This diagnostic bottleneck is fundamentally driven by a conflict between high clinical complexity and low epidemiological prevalence: rare phenotypes are often highly heterogeneous and partially penetrant, and they mimic common disorders, resulting in presentations that challenge the limits of any individual clinician's experience\cite{taruscio_rare_2024,stoller_challenge_2018}. While knowledge bases such as OMIM\cite{hamosh_online_2021} and Orphanet (ORPHA)\cite{weinreich_orphanet_2008} provide essential standardized vocabularies and phenotypic profiles, the broader clinical evidence required for accurate diagnosis remains fragmented across unstructured literature and disparate registries. Consequently, a structural diagnostic gap persists even within elite tertiary referral centres, where specialized clinical expertise is frequently stretched to its limits\cite{michaels-igbokwe_standardized_2021}.

The rapid development of artificial intelligence (AI) has led to the introduction of successive computational strategies to address this diagnostic gap\cite{chen_rarebench_2024}. Early efforts focused on knowledge-encoded symbolic reasoning, in which ontology-driven tools such as Phenomizer\cite{kohler_clinical_2009} leveraged Human Phenotype Ontology (HPO)\cite{gargano_human_2024} annotations formalize disease--phenotype matching with mathematical rigor that transcended unaided intuition. The subsequent application of deep learning expanded these capabilities to high-dimensional clinical and imaging modalities, enabling automated detection of syndromic facial dysmorphic features\cite{gurovich_identifying_2019}, retinal anomalies\cite{pontikos_next-generation_2025}, and complex radiographic patterns\cite{flynn_deep_2023}. More recently, large language models (LLMs) have increased the diagnostic ceiling by leveraging their unique capacity for contextual reasoning and generative synthesis to integrate clinical notes, multiomic datasets, and medical literature\cite{yang_rdguru_2025,zhao_agentic_2026}. This evolution has led to the development of autonomous agentic architectures. By coupling iterative hypothesis generation with dynamic information retrieval and multistep reasoning, these agentic pipelines have achieved high diagnostic accuracy when processing curated patient vignettes, occasionally matching the performance of specialized clinical teams\cite{zhao_agentic_2026}.

Despite these advancements in various AI paradigm, to exploit the synergy of their complementary strengths thus far largely unrealized.  While knowledge-driven symbolic models offer deductive rigor and auditable reasoning pathways by leveraging structured clinical ontologies\cite{kohler_clinical_2009,bauer_bayesian_2012}, neural architectures, including deep-learning models and generative LLMs, provide inductive power and heuristic flexibility through high-dimensional pattern recognition. Although neuro-symbolic AI frameworks attempt to bridge this divide by combining these orthogonal approaches\cite{prenosil_neuro-symbolic_2025}, their application to RD diagnosis is still limited. Recent systems have attempted superficial aggregation; for instance, the deep-learning ensemble model PhenoBrain utilizes a multilayer perceptron (MLP) classifier with symbolic computing\cite{mao_phenotype-based_2025}, the agentic system RDguru applies reinforcement learning to align recommendations from multiple diagnostic tools\cite{yang_rdguru_2025}, and DeepRare relies on agentic reflection based on retrieved evidence and outputs from diverse diagnostic tools\cite{zhao_agentic_2026}. However, these integration strategies typically aggregate outputs downstream using linear weighting or contextual prompting, rather than combining the underlying architectures. Reconciling the discrete, deterministic logic of clinical ontologies with the fluid, heuristic reasoning of autonomous agents---and other valid paradigms---represents an unresolved methodological challenge.

At the same time, the translation of AI-assisted diagnostics into real-world clinical workflows remains constrained by systematic evaluation artefacts that often result in the overestimation of model capabilities\cite{Goldenberg2026}. Three primary validation biases frequently decouple benchmark performance from clinical utility: (i) data contamination, wherein evaluation cases overlap with the information in  pretraining datasets or development corpora\cite{cheng_survey_2025}; (ii) contextual enrichment or hindsight bias, in which retrospective vignettes incorporate post-hoc diagnostic cues and specialist framings unavailable during \textit{de novo} clinical encounters\cite{ghassemi_false_2021}; and (iii) documentation-style bias, where models exploit institutional linguistic idiosyncrasies and recording tendencies as diagnostic shortcuts rather than performing generalized medical reasoning\cite{zech_variable_2018}. These vulnerabilities aggregate to cause  ``in-distribution inflation'', masking the performance degradation of models when applied to prospective, uncurated, or clinically heterogeneous datasets. This inflation is particularly problematic in RD diagnosis, where the vast differential space of more than 7,000 conditions amplifies the gap between reported benchmark accuracy and actual diagnostic resolution. Mitigating these biases requires a rigorous evaluation framework---one anchored in temporal or even institutional isolation---to stress test models with independent clinical datasets that preserve the real-world ambiguity of uncurated clinical practices.

Here, we introduce \textbf{FusionRare}, a unified framework that architecturally reconciles symbolic deductive reasoning with neural inductive learning for RD diagnosis. FusionRare integrates four complementary diagnostic paradigms---knowledge-driven computational reasoning, deep-learning pattern recognition, similar-case retrieval, and heuristic LLM inference---through a statistical meta-aggregation layer to establish a deterministic boundary and is subsequently refined via a knowledge-enhanced agentic reflection step.
To enhance the clinical interpretability of diagnostic recommendations, we also present \textbf{RDmap}---a two-dimensional navigable manifold generated through semantic phenotype similarity mapping. RDmap projects discrete disease entities onto a continuous low-dimensional coordinate space, translating abstract clinical priorities into an intuitive topological interface.
We evaluated FusionRare using rigorous antileakage protocols on 2,528 RD cases across public registries and two independent clinical centres.
In diverse testing scenarios, FusionRare consistently outperformed existing baselines. In phenotype-only tasks, the system achieved a macroaverage Recall@1 of 51\%, representing a 19\% absolute improvement over current state-of-the-art (SOTA) diagnostic systems, while maintaining stable performance across medical specialties and institutional boundaries. 
When phenotypic presentation data were combined with genomic variant data, FusionRare achieved a disease prioritization Recall@1 of 79\%, substantially exceeding that of specialized pipelines such as Exomiser (48\%)\cite{smedley_next-generation_2015} and LIRICAL (68\%)\cite{robinson_interpretable_2020}.
A clinical trial involving 20 multispecialty physicians demonstrated the utility of the framework: using FusionRare elevated the rank of the true diagnosis in 45\% of decisions, improved the subjective confidence of physicians in 62\% of decisions, and achieved a 21\% reduction in time to diagnosis. Collectively, our results suggest that the synergistic integration of heterogeneous paradigms can effectively mitigate systematic biases and clinical uncertainty in the diagnosis of complex, rare conditions.

\section{Results}\label{sec2}

\begin{figure}[t]
    \centering
    \includegraphics[width=1.0\textwidth]{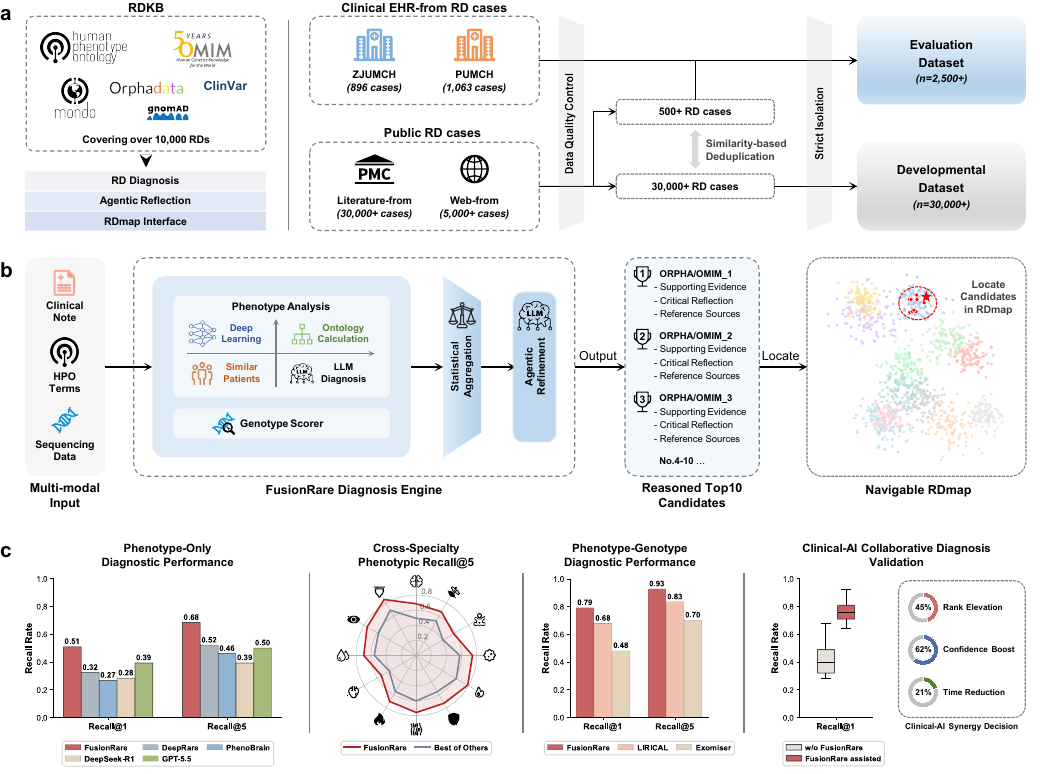} 
    \caption{\textbf{Study overview.} \textbf{(a)} Data foundation. Construction of the Rare Disease Knowledge Base (RDKB) and Patient Bank under rigorous antileakage protocols, ensured strict isolation between developmental and evaluation datasets. \textbf{(b)} FusionRare diagnosis engine. The multiparadigm pipeline integrates four orthogonal reasoning modules whose outputs are synthesized via statistical aggregation and refined by agentic LLM reflection. The pipeline yields prioritized diagnostic candidates and their spatial localization on the navigable RDmap. \textbf{(c)} Performance benchmarking and clinical validation. Evaluation of FusionRare across phenotype-only, cross-specialty, phenotype-genotype, and clinical application settings, revealed that the performance of FusionRare was consistently superior to that of SOTA methods and that it has tangible clinical utility.
    \label{fig:study_overview}}
\end{figure}

\subsection{System Overview}

FusionRare is structured as a hierarchical system through a multistage funnel framework (Fig.~\ref{fig:study_overview}b). The initial stage includes four orthogonal reasoning paradigms that are run in parallel to generate heterogeneous diagnostic signals: (i) knowledge- and ontology-driven Bayesian computing, (ii) deep-learning pattern recognition via a residual network, (iii) case-based similarity retrieval querying a dynamic repository of more than 30,000 RD cases, and (iv) zero-shot heuristic LLM inference. Rather than relying on multistep planning and iterative retrieval, which can introduce high variance in autonomous agentic systems, FusionRare supports downstream reasoning within a robust statistical consensus. In the intermediate stage, these diverse source signals are integrated through a statistical meta-aggregation layer to generate a prioritized consensus ranking, establishing a deterministic and mathematically auditable boundary. In the final refinement stage, this consensus is processed by an agentic reflection module using chain-of-thought (CoT)\cite{wei_chain--thought_2022} reasoning. This module resolves conflicts among initial diagnostic engines by integrating structured clinical priors with retrieved medical evidence and balancing mathematical rigor with heuristic flexibility.

FusionRare covers 10,260 RD entities aligned with OMIM and ORPHA nomenclatures and anchored to a synchronized 2023 Q1 knowledge snapshot. Designed to accommodate multimodal clinical workflows, the framework supports both phenotype-only prioritization and phenotype-genotype joint diagnosis, allowing clinicians to integrate structured HPO profiles with genomic sequencing data (in VCF format) to identify candidate diseases and their causative genetic variants. To support clinical decision-making, FusionRare projects the high-dimensional phenotypic relationships among RDs onto RDmap---a two-dimensional semantic manifold serving as a topological reference. By positioning a patient's clinical profile proximal to phenotypically related disease clusters, RDmap translates abstract diagnostic ranks into a navigable coordinate space, enabling clinicians to visually explore and evaluate diagnostic hypotheses across the entire RD spectrum.

\subsection{Experimental Framework and Benchmarking Design}\label{benchmarking}

We developed and evaluated FusionRare using a multitiered benchmarking framework comprising more than 40,000 RD cases (Extended Data Table~\ref{tab:case_sources_final}). To distinguish prospective reasoning capabilities from the system's ``empirical memory'', we implemented a tripartite data hierarchy. \textbf{Tier A} and \textbf{Tier B} cohorts, totalling 38,729 cases from literature-linked datasets (e.g., RareArena\cite{chen_rarearena_2026}, Phenopacket Store\cite{danis_corpus_2025}) and global registries (e.g., DECIPHER\cite{firth_decipher_2009}, RAMEDIS\cite{topel_ramedis_2006}), were predominantly allocated for development, with a small subset reserved for testing. Most of these cases were linked to PubMed Identifiers (PMIDs) and curated phenotype records, providing training data for the deep-learning classifier and populating the case retrieval index. In contrast, \textbf{Tier C} serves as the independent clinical validation core, consisting of more than 2,000 strictly isolated cases from two medical centres: the Children's Hospital of Zhejiang University School of Medicine (ZJUMCH) and the Peking Union Medical College Hospital (PUMCH). The ZJUMCH cases were processed via internal pipelines with manual curation, whereas the PUMCH cases were automatically extracted from electronic health records (EHRs). The PUMCH cohort exhibited high phenotypic density (averaging 31.2 HPO terms per case), reflecting the clinical noise and redundancy typical of automated EHR extraction. This multicentre dataset, spanning more than 20 medical specialties and varying in phenotypic density, provides a rigorous test of model robustness across diverse, real-world clinical presentations.

We benchmarked FusionRare against representative diagnostic models spanning multiple computational paradigms: (i) \textbf{knowledge-driven computational models} (e.g., LIRICAL\cite{robinson_interpretable_2020}, PhenoPro\cite{li_phenopro_2019}, BOQA\cite{bauer_bayesian_2012}); (ii) \textbf{deep-learning-based ensemble models} (e.g., PhenoBrain\cite{mao_phenotype-based_2025}, PhenoDP\cite{wen_phenodp_2025}); (iii) \textbf{autonomous agentic systems} (DeepRare\cite{zhao_agentic_2026}); and (iv) \textbf{generative LLMs} (e.g., GPT-4o\cite{openai_gpt-4_2024}, DeepSeek-V3\cite{deepseek-ai_deepseek-v3_2025}, the reasoning model DeepSeek-R1\cite{guo_deepseek-r1_2025}, and the frontier LLM GPT-5.5\cite{openai_introducing_2026}). For phenotype-genotype joint analysis, FusionRare was compared with established multimodal pipelines, specifically Exomiser\cite{smedley_next-generation_2015} and LIRICAL\cite{robinson_interpretable_2020}.

To control for systemic evaluation bias, we instituted three rigorous safeguard protocols. (i) \textbf{Temporal Knowledge Alignment:} We synchronized our internal rare disease knowledge base (RDKB) and all local baseline implementations to a 2023 Q1 knowledge snapshot (encompassing OMIM, Orphanet, and background databases for Exomiser/LIRICAL). This alignment prevents performance inflation driven by post-2023 medical discoveries and ensures that all evaluated models operate under identical knowledge constraints. (ii) \textbf{AntiLeakage and Data Isolation:} In addition to evaluating the framework using institutional cohorts (Tier C), we enforced a stringent Jaccard similarity filter (threshold $<0.4$) and PMID-level exclusion for all literature-derived test cases. This safeguard ensured that the benchmarks measured generalized diagnostic reasoning rather than pattern memorization. (iii) \textbf{Foundation LLM Parity:} To isolate the architectural contributions of FusionRare from the baseline capabilities of the underlying LLMs, we restricted our primary framework to DeepSeek-V3 and DeepSeek-R1, matching the LLM backbone reasoning performance of comparative agents such as DeepRare (which relies on DeepSeek‑V3 and GPT‑4o). More advanced frontier models---such as GPT-5.5---were confined to comparative testing and ablation studies.

\subsection{Comprehensive Phenotypic Diagnostic Performance}\label{full-cohort}

\begin{figure}[ht!]
    \centering
    \includegraphics[width=1.0\textwidth]{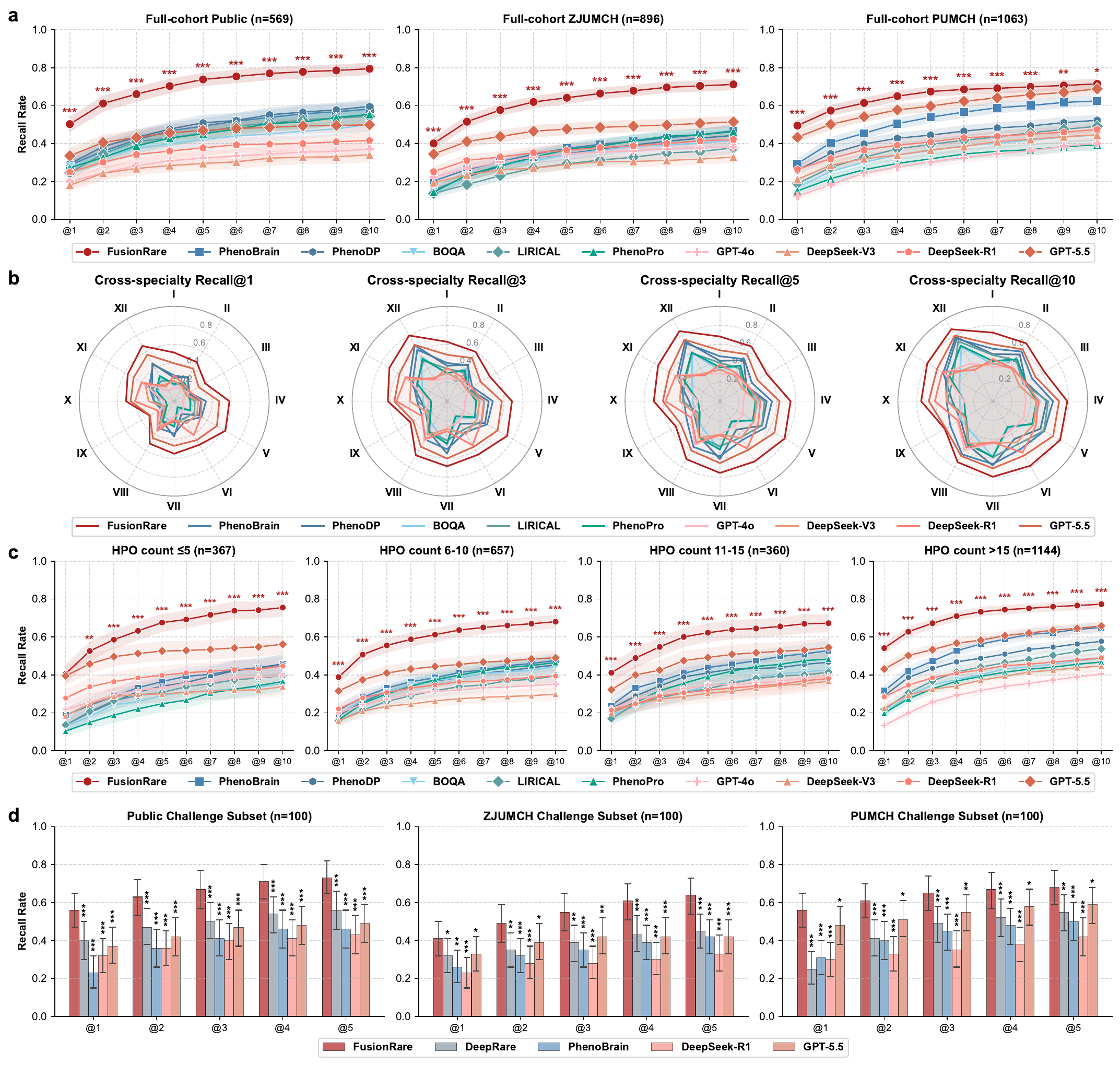} 
    \caption{\textbf{Diagnostic performance across datasets, specialties, and complexities.} \textbf{(a)} Benchmarking on three independent cohorts. \textbf{(b)} Performance across medical specialties. Radar charts illustrating Recall@$k$ at varying ranks ($k=1, 3, 5, 10$) across 12 major disease categories: I, Nervous ($n$=1442); II, Musculoskeletal (828); III, Development/Morphogenesis (765); IV, Cancer/Tumour (740); V, Endocrine (681); VI, Immune (613); VII, Connective tissue (598); VIII, Metabolic (591); IX, Cardiovascular (465); X, Haematological (396); XI, Visual (380); and XII, Reproductive (355). \textbf{(c)} Impact of phenotypic complexity. Diagnostic recall is stratified by the number of HPO terms per case. \textbf{(d)} Clinical consistency evaluation of FusionRare against SOTA systems (DeepRare, GPT-5.5, PhenoBrain, and DeepSeek-R1) using manually curated challenge subsets. \textbf{Figure Notes:} Data in (a, c, d) are presented as the mean $\pm$ 95\%CI. Statistical significance was determined using a one-sided bootstrap test: *$P<0.05$, **$P<0.01$, ***$P<0.001$.}
    \label{fig:full-cohort}
\end{figure}

We evaluated the phenotypic diagnostic performance of FusionRare by conducting a large-scale comparative benchmark using 2,528 strictly isolated cases. FusionRare consistently outperformed specialized deep-learning models, Bayesian models, and frontier LLMs (Fig.~\ref{fig:full-cohort}a), demonstrating statistically significant improvements compared with the baseline models ($P<0.05$) across all cohorts.

For the public cohort, FusionRare achieved a Recall@1 of 50.3\% (95\% CI: 46.2--54.5\%), whereas DeepSeek-R1, PhenoDP, PhenoBrain, and the frontier model GPT-5.5 achieved Recall@1 values of 25.0\%, 29.4\%, 28.6\%, and 33.6\%, respectively. This advantage was maintained in real-world clinical datasets despite the inherent complexity of EHRs. For the ZJUMCH cohort, FusionRare achieved a Recall@1 of 40.1\% (95\% CI: 36.9--43.1\%), outperforming GPT-5.5 (34.5\%). For the high-density PUMCH cohort, the Recall@1 of the framework was 49.5\% (95\% CI: 46.5--52.7\%), surpassing that the values of both the strongest deep-learning baseline, PhenoBrain (29.4\%), and GPT-5.5 (43.4\%). Notably, while Recall@1 showed cohort-dependent variations, FusionRare's Recall@10 remained stable across all the datasets (ranging from approximately 71.2\% to 79.4\%). This consistent performance suggests that integrating multiple diagnostic paradigms stabilizes the prioritized differential list within a narrow clinical window, even when highly divergent EHR data are processed.

The generalizability of FusionRare was further evaluated across different medical specialties. Stratification into 12 major disease categories---encompassing the majority of rare conditions---confirmed a consistent advantage across all nosological domains (Fig.~\ref{fig:full-cohort}b). We also analysed the system's sensitivity to varying clinical information density (Fig.~\ref{fig:full-cohort}c). In information-scarce scenarios (defined by an HPO count $\leq$ 5), which correspond to early-stage clinical encounters, FusionRare maintained a Recall@1 of 40.6\% (95\% CI: 36.0--45.8\%), whereas PhenoBrain and DeepSeek-R1 achieved Recall@1 values of 18.8\% and 27.8\%, respectively. Although FusionRare's Recall@1 was comparable to that of GPT-5.5 (39.5\%) in these sparse cases, it demonstrated a statistically significant advantage from Recall@2 onward. As clinical detail increased (HPO count $>$ 15), FusionRare's performance scaled to a Recall@1 of 54.1\%. The performance margin between FusionRare and the next-best model remained statistically significant across all complexity bins (typically $P<0.001$), demonstrating that the multiparadigm framework approach is adaptable to diverse scenarios ranging from sparse phenotypic presentations to high-complexity clinical profiles.

Owing to the current limited-access volume of the agent-based SOTA method DeepRare, we conducted a comparative evaluation using a manually curated challenge subset ($n=300$), comprising an equal number of cases randomly selected from the Public, ZJUMCH, and PUMCH cohorts. Considering that the exact matching of database identifiers cannot truly reflect the complex scenarios in multiple subtypes of diseases, we adopted a clinical consistency assessment for this evaluation. Two authors independently audited the generated differential diagnoses, classifying a recommendation as correct if it captured synonymous conditions, relevant clinical subtypes, or accurate higher-order disease categories, and as incorrect if it contained fundamental clinical conflicts (e.g., incorrect causative genes or mutually exclusive subtypes). We compared FusionRare to the agentic diagnostic system DeepRare and the frontier model GPT-5.5, as well as the deep-learning ensemble model PhenoBrain and the reasoning model DeepSeek-R1.

On the basis of the challenge subsets, FusionRare outperformed both DeepRare and GPT-5.5 across all evaluated sources (Fig.~\ref{fig:full-cohort}d). For the Public Challenge Subset, FusionRare achieved a Recall@1 of 56.0\% (95\% CI: 46.0--65.0\%), compared with 40.0\% for DeepRare ($P<0.001$) and 37.0\% for GPT-5.5 ($P<0.001$). At higher ranks, FusionRare achieved a Recall@5 of 73.0\% (95\% CI: 64.0--81.0\%), outperforming DeepRare (56.0\%, $P<0.001$) and GPT-5.5 (48.0\%, $P<0.001$). Similar trends were observed for the institutional cohorts. For the ZJUMCH cohort, the difference in Recall@1 was smaller, although still statistically significant (41.0\% for FusionRare versus 32.0\% for DeepRare and 33.0\% for GPT-5.5; $P<0.05$). However, this margin expanded for Recall@3 (55.0\% versus 39.0\% and 42.0\%, respectively; $P<0.01$) and Recall@5 (64.0\% versus 45.0\% and 42.0\%, respectively; $P<0.001$). For the high-density PUMCH cohort, FusionRare achieved a Recall@1 of 56.0\%, outperforming DeepRare (25.0\%, $P<0.001$) and GPT-5.5 (48.0\%, $P<0.05$). Although DeepRare performed better than the deep-learning baseline PhenoBrain at ranks greater than Recall@2, its performance was inferior to that of both FusionRare and GPT-5.5. These observations suggest that end-to-end agentic systems such as DeepRare can be sensitive to the clinical noise and redundancy typical of automated EHR extraction, whereas FusionRare's multiparadigm fusion architecture effectively identifies the diagnostic core features. The results indicate that integrating heterogeneous computational paradigms provides a more stable clinical recommendation model than relying solely on autonomous agents or general LLMs does. This advantage is complemented by a marked improvement in diagnostic efficiency: DeepRare averaged over ten minutes per case, compared with approximately one minute for FusionRare. It will also save a large amount of token consumption.

\subsection{Integrated Phenotypic-Genomic Diagnostic Performance}

Given that majority of RDs have a genetic aetiology, establishing a definitive diagnosis requires the integration of deep phenotypic characterization with molecular evidence. To evaluate the multimodal performance of FusionRare, we compiled a multimodal validation cohort ($n=258$) on the basis of the isolated public cohort encompassing 164 diseases and 153 causative genes, sourced from the LIRICAL ($n=190$), MME ($n=36$), and VPI ($n=32$) datasets. To replicate the clinical challenge in which rare pathogenic variants are obscured by an abundance of variants of uncertain significance, we employed a stochastic genomic perturbation strategy. This involved spiking validated pathogenic mutations into ten independent whole-exome (WES) backgrounds randomly selected from an in-house repository of real-world exomes, each containing approximately 40,000 single nucleotide polymorphisms (SNPs) or insertions/deletions (indels). This spike-in methodology is widely used for benchmarking gene diagnostic systems\cite{smedley_next-generation_2015,robinson_interpretable_2020}. Our stochastic genomic perturbation approach serves as a robust stress test to evaluate the system's ability to isolate true pathogenic signals across diverse genetic backgrounds rather than within a static genomic context.

\begin{figure}[t!]
    \centering
    \includegraphics[width=1.0\textwidth]{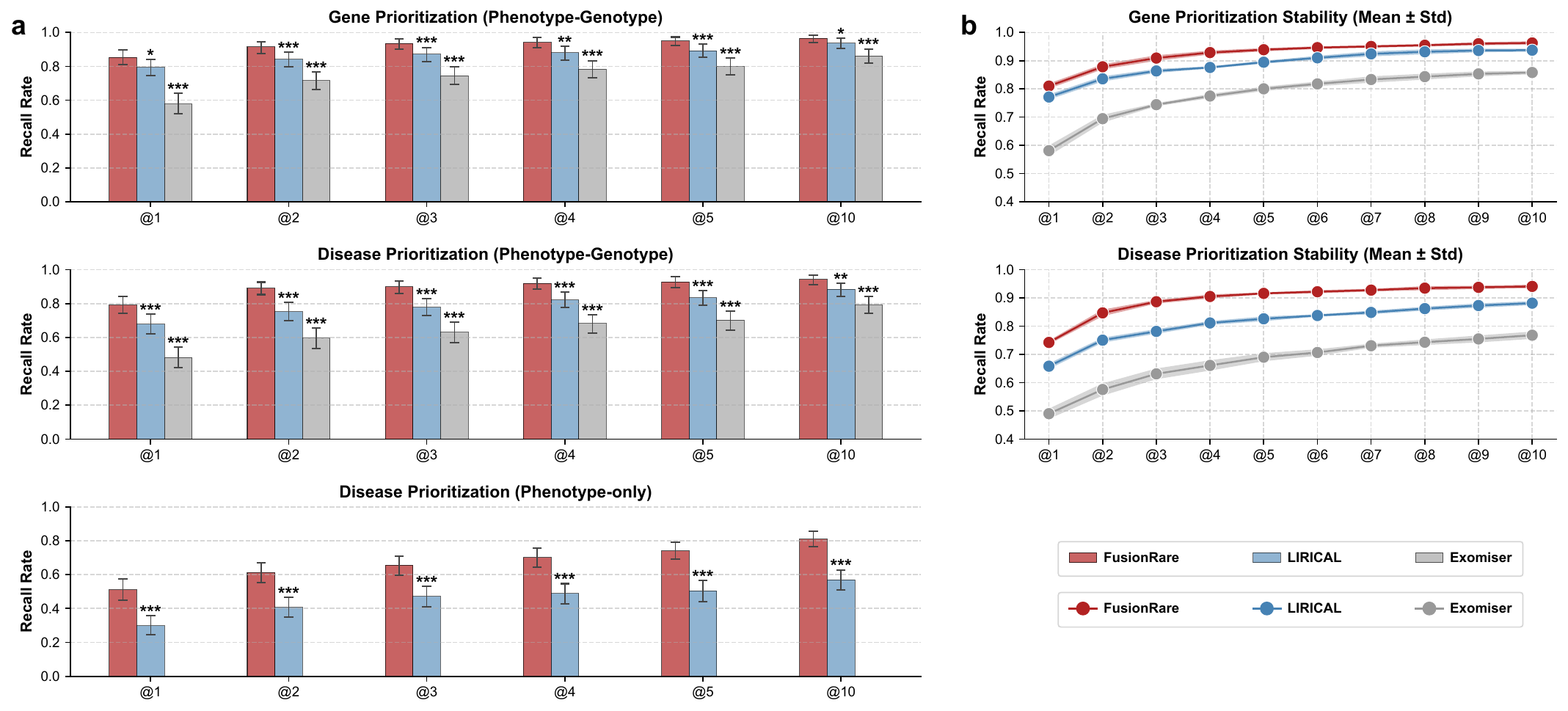} 
    \caption{\textbf{Integrated phenomic-genomic diagnostic performance and stability analysis}. \textbf{(a)} Comparative benchmarking. Recall@$k$ for gene prioritization (top), disease prioritization (middle), and phenotype-only disease prioritization (bottom). \textbf{(b)} Diagnostic stability. Performance curves showing the mean $\pm$ Std across ten independent trials with heterogeneous genomic backgrounds. \textbf{Figure Notes:} Data in (a) are presented as the mean $\pm$ 95\%CI; Shaded areas in (b) denote Std; Statistical significance was determined using a one-sided bootstrap test: *$P<0.05$, **$P<0.01$, ***$P<0.001$.}
    \label{fig:gene_test}
\end{figure}

Statistical analysis of the median consensus ranks confirmed the diagnostic accuracy of the framework in the joint phenotype-genotype analysis (Fig.~\ref{fig:gene_test}a). In disease prioritization, FusionRare achieved a Recall@1 of 79.1\% (95\% CI: 74.0--84.1\%), outperforming LIRICAL (67.8\%, $P<0.001$) and Exomiser (48.1\%, $P<0.001$). The corresponding Recall@3 was 89.9\%, compared with 77.9\% for LIRICAL and 63.2\% for Exomiser. For gene prioritization, FusionRare identified the causative gene as the primary candidate in 85.3\% of cases (95\% CI: 80.6--89.5\%), outperforming LIRICAL (79.5\%) and Exomiser (57.8\%). Across all ranks, FusionRare achieved a gene Recall@3 of 93.4\%, providing resolved candidate lists for clinical genomic interpretation. Furthermore, FusionRare was resilient to genomic background variation. In the multitrial stability evaluations, the framework achieved the highest mean performance and the lowest stochastic variance (Fig.~\ref{fig:gene_test}b). Specifically, the standard deviation ($\sigma$) of disease prioritization for the Recall@1 and Recall@5 of FusionRare remained low ($0.67\%$ and $0.37\%$, respectively) compared with those of LIRICAL ($\sigma = 0.96\%$ and $0.83\%$, respectively) and the baseline Exomiser ($\sigma = 1.80\%$ and $1.43\%$, respectively). This consistent performance indicates that the framework maintains diagnostic precision despite the biological noise introduced by individual-specific genomic backgrounds.

\subsection{Randomized Crossover Clinician-AI Diagnosis Validation}
\begin{figure}[ht!]
    \centering
    \includegraphics[width=1.0\textwidth]{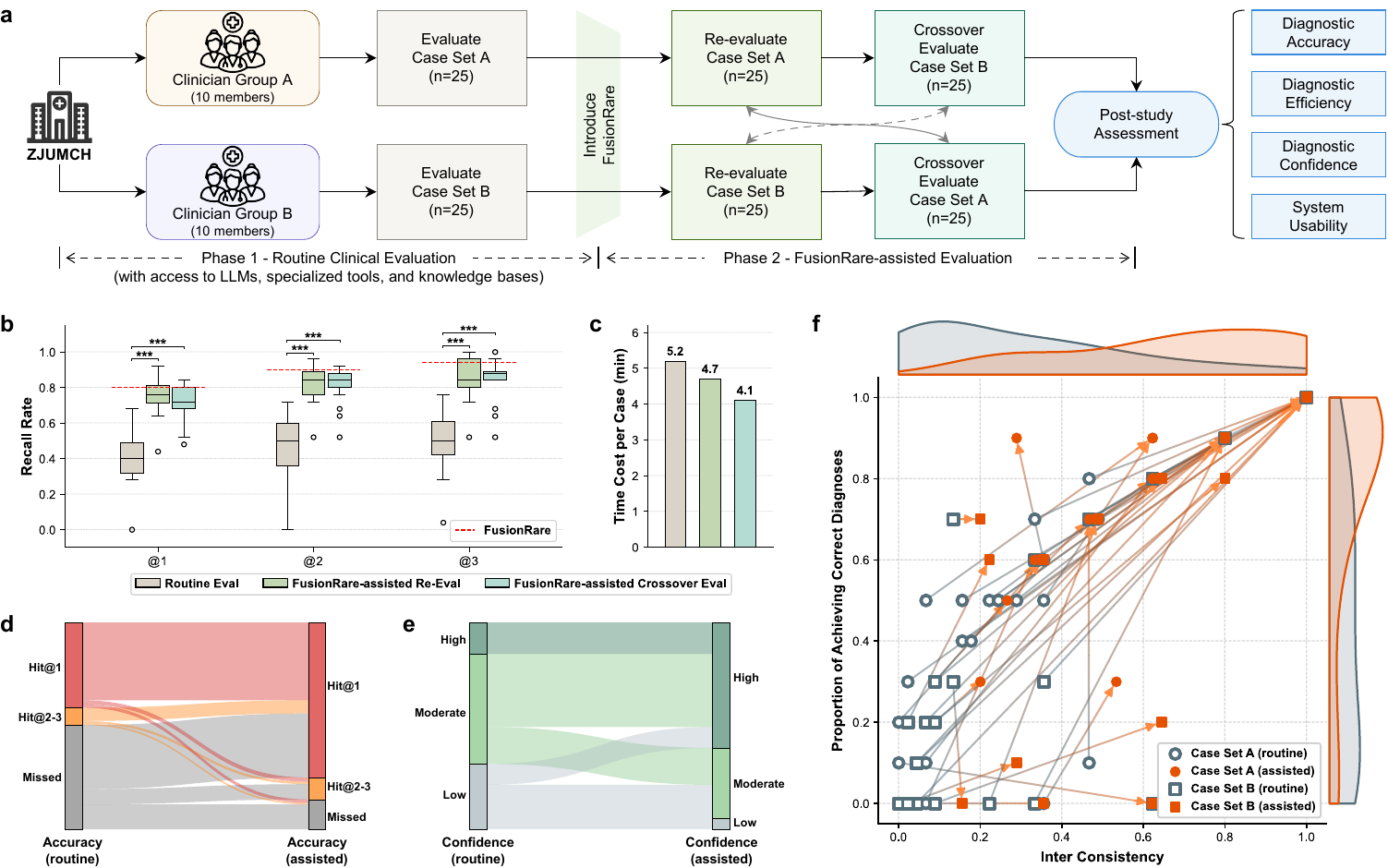} 
    \caption{\textbf{Clinical validation of FusionRare.} \textbf{(a)} Multispecialty randomized crossover clinical trial design. \textbf{(b)} Diagnostic accuracy of clinicians across three evaluation stages. Accuracy is represented as the macroaverage Recall@$n$ ($n=1,2,3$) across individual clinicians for initial routine evaluation, re-evaluation with FusionRare, and crossover evaluation of \textsl{de novo} cases with FusionRare in the form of box plots. Box plots show the median (centreline), upper and lower quartiles (box limits), and whiskers ($1.5\times$interquartile range). The horizontal dashed lines indicate the performance of FusionRare. \textbf{(c)} Clinician diagnostic efficiency. The bars represent the mean time to diagnosis per case. \textbf{(d, e)} Trajectories of diagnostic refinement and confidence modulation. Sankey diagrams illustrate the flow of 500 paired decisions from initial routine assessment to FusionRare-assisted re-evaluation. Red flows (d) highlight the restratification of diagnostic ranks, while green flows (e) depict the evolution of subjective diagnostic confidence. \textbf{(f)} Case-level harmonization of diagnostic accuracy and consensus. The scatter\&vector plot visualizes the topological shift of 50 individual cases from routine diagnosis (empty markers) to FusionRare-assisted diagnosis (filled markers) states (Y-axis: proportion of correct clinician diagnoses; X-axis: pairwise agreement rate among clinician diagnoses). Marginal density curves illustrate the score distributions of the correct diagnosis rate (right) and consistency (upper) among clinicians operating in routine (blue curves) versus assisted (orange curves) diagnostic settings. \textbf{Figure Notes:} In (b), the statistical significance between re-evaluation and initial evaluation was assessed using the Wilcoxon signed-rank test, whereas the significance between crossover evaluation and initial evaluation was assessed using the Mann-Whitney U test. *$P<0.05$, **$P<0.01$, ***$P<0.001$.}
    \label{fig:trial}
\end{figure}

To evaluate the clinical utility and human-AI collaborative potential of FusionRare, we conducted a clinical validation study (Fig.~\ref{fig:trial}a) involving 20 clinicians of varying seniority. Clinicians were recruited from diverse departments, including Neonatology, Genetics and Metabolic Disorders, Immunology, Cardiology, Growth and Development, Endocrinology, and the Intensive Care Unit (ICU). The study utilized 50 curated cases from evaluation cohorts, each representing a unique rare disorder, drawn to be broadly representative yet sufficiently challenging. The trial was structured into two sequential phases. In \textbf{Phase 1 (Routine Clinical Evaluation)}, clinicians were randomized into two groups to evaluate distinct case sets ($n=25$ each) using their standard diagnostic resources. These resources included unrestricted access to medical databases, specialized bioinformatic tools, and general LLM chatbots (e.g., DeepSeek, ChatGPT, and DouBao), resulting in a standard-of-care baseline. Clinicians provided up to three differential diagnoses for each case. In \textbf{Phase 2 (FusionRare-assisted Evaluation)}, following an orientation to the FusionRare platform, clinicians re-evaluated their assigned cases with system assistance to measure shifts in diagnostic accuracy and subjective confidence. To evaluate generalizability and control for memory bias, we implemented a crossover design wherein each clinician group evaluated the remaining 25 novel, previously unseen cases with the assistance of FusionRare. Clinicians remained blinded to the ground-truth diagnoses throughout both phases. We also performed a multidimensional assessment of the human--AI interface, measuring diagnostic efficiency (time to diagnosis), cognitive load via the NASA Task Load Index (NASA-TLX), and system usability via the System Usability Scale (SUS). Detailed questionnaires are provided in Appendix~D. The diagnostic outcomes and scale scores were compiled for statistical analysis.

FusionRare significantly improved the diagnostic accuracy of clinicians while maintaining robust generalizability across case sets. During the initial routine evaluation phase, the clinicians achieved a median Recall@1 of 40.0\% (Fig.~\ref{fig:trial}b). With the assistance of FusionRare, this performance increased to 76.0\% ($P<0.001$, Cohen's $d_z=3.02$). Notably, the diagnostic performance in the crossover phase was comparable to that observed during the re-evaluation of previously seen cases, suggesting that the system successfully guided clinicians through unfamiliar clinical scenarios without the introduction of memory bias. Furthermore, approximately 25\% of the clinician-AI collaborative sessions achieved higher diagnostic accuracy than the standalone FusionRare system did, indicating synergistic integration, where expert clinical reasoning and model recommendations complement each other. Granular trajectory analysis revealed diagnostic refinement with minimal evidence of automation bias. FusionRare-assisted evaluation led to diagnostic rank improvements in 44.8\% of decisions, whereas a decline in rank occurred in only 4.4\% of decisions, resulting in a rectification-to-misleading odds ratio of 10.2 (Fig.~\ref{fig:trial}d). This positive clinical outcome was accompanied by an increase in subjective clinician confidence in 61.9\% of diagnostic sessions (Fig.~\ref{fig:trial}e). Case-level analysis also revealed topological convergence towards correct diagnoses: the use of FusionRare improved individual accuracy and significantly increased interrater consistency (Fig.~\ref{fig:trial}f), narrowing the performance gap between clinicians with varying experience levels. Finally, FusionRare improved diagnostic efficiency and reduced cognitive workload. Compared with routine clinical practice, assisted diagnostic sessions achieved a 21.5\% reduction in time to diagnosis (Fig.~\ref{fig:trial}c). Subjective workload assessment via the NASA-TLX confirmed a low level of cognitive burden associated with the platform, with clinicians reporting low mean scores for mental demand (39/100), temporal demand (28/100), physical demand (20/100), and frustration (20/100). A mean SUS score of 88/100 indicates high usability, suggesting that FusionRare can serve as a supportive clinical decision support system to help alleviate the diagnostic and cognitive burdens associated with RD management.

\subsection{Ablation Study and Mechanistic Analysis}\label{ablation-study}

\begin{figure}[ht!]
    \centering
    \includegraphics[width=1.0\textwidth]{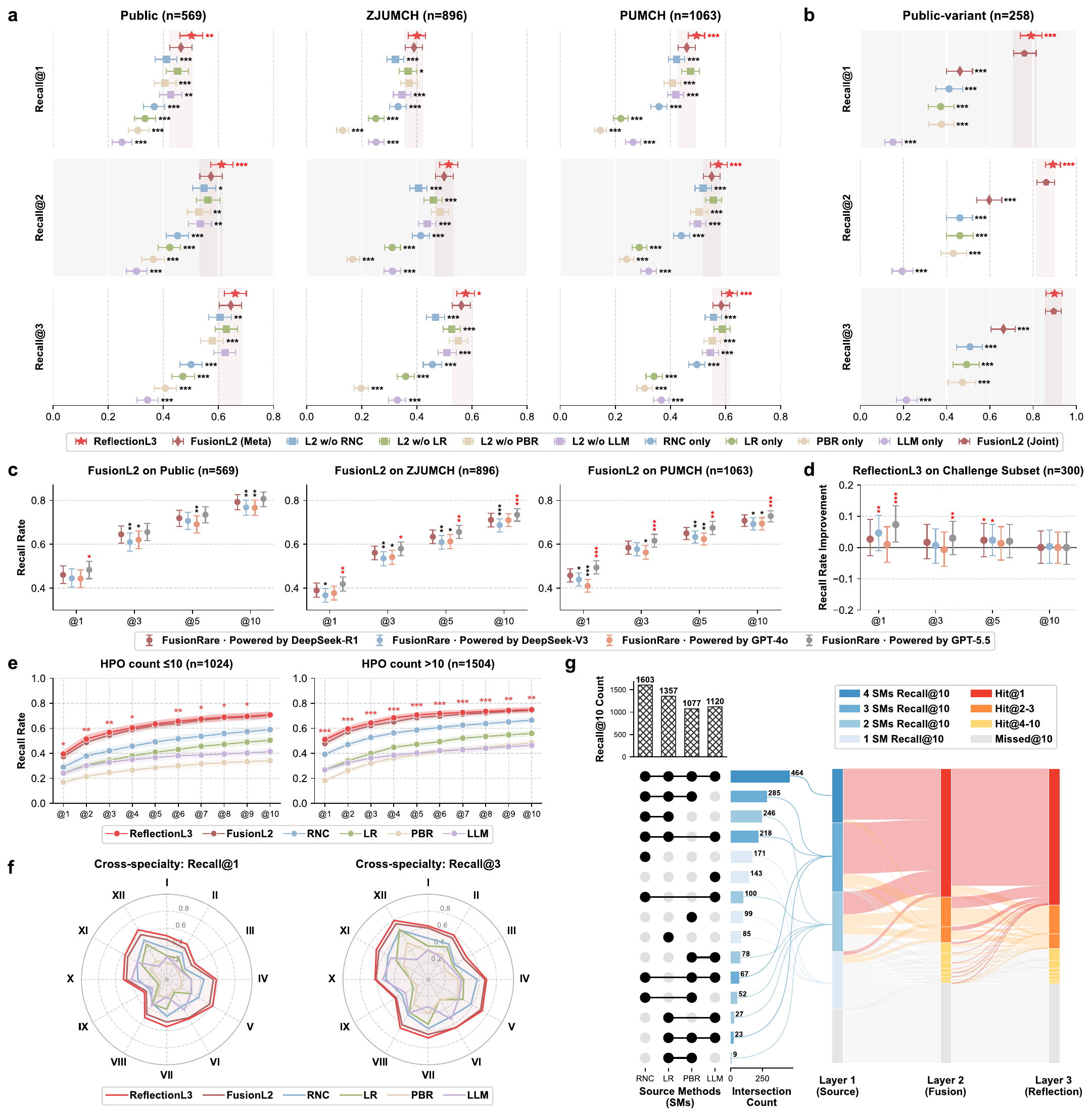} 
    \caption{\textbf{Ablation study and mechanistic insights into FusionRare.} \textbf{(a)} Synergy of source paradigms. Comparison of diagnostic recall across three cohorts for the full pipeline: source paradigms, statistical aggregation (FusionL2), agentic refinement (ReflectionL3), and ablated versions of FusionL2. \textbf{(b)} Genomic enhancement. Significant performance improvement provided by the FusionL2 (Joint) model for the public-variant cohort. \textbf{(c, d)} Foundation LLM sensitivity. Effects of the underlying LLM backbones (DeepSeek-R1, DeepSeek-V3, GPT-4o, and GPT-5.5) on FusionL2 performance (c) and absolute recall gains (versus the FusionL2 baseline) of ReflectionL3 (d). \textbf{(e, f)} Robustness across complexities and specialties. Performance stability of the fusion hierarchy stratified by HPO term count (e) and across 12 medical specialties (f). \textbf{(g)} Evolution of the ``Fusion'' workflow. Upset plot (left) showing the intersection of successful diagnoses among source paradigms, and a Sankey diagram (right) illustrating the stepwise promotion of true diagnoses from Round 1 (source paradigms) through Round 2 (statistical aggregation) to Round 3 (agentic reflection). \textbf{Figure Notes:} Data in (a-e) are presented as the mean $\pm$ 95\%CI; statistical significance was determined based on a one-sided bootstrap test: *$P<0.05$, **$P<0.01$, ***$P<0.001$; statistical significance is assessed against FusionL2 (Meta) for (a), FusionL2 (Joint) for (b), FusionL2 powered by DeepSeek-R1 for (c), FusionL2 for (d), and ReflectionL3 for (e); statistically significant markers: black markers indicate significantly lower performance than baseline, while red indicates significantly higher performance; the disease category labels in (f) correspond to those in Fig.~\ref{fig:full-cohort}b.}
    \label{fig:ablation}
\end{figure}

To elucidate the individual contributions of each diagnostic paradigm and the efficacy of our hierarchical fusion architecture, we performed a series of rigorous ablation studies and mechanistic analyses (Fig.~\ref{fig:ablation}). Corresponding to our multistage funnel framework shown in Fig.~\ref{fig:study_overview}b, the FusionRare diagnosis pipeline consists of three sequential layers: (i) \textbf{Source Paradigms}, comprising the Residual Network Classifier (RNC), Likelihood Ratio (LR) model, Patient Bank Retriever (PBR), and LLM-based zero-shot diagnosis; (ii) \textbf{Statistical Aggregation (FusionL2)}, with ``Meta'' (phenotype-only) and ``Joint'' (phenotype-genotype) variants; and (iii) \textbf{Agentic Refinement (ReflectionL3)}.

\textbf{Paradigm Orthogonality and Source Model Necessity.}
The robustness of FusionRare arises from the complementarity between its diagnostic engines (Fig.~\ref{fig:ablation}a). While the deep-learning (RNC) and heuristic-reasoning (LLM) modules were critical across all the datasets, the knowledge-driven (LR) and case-retrieval (PBR) models exhibited context-specific strengths. The LR engine was particularly effective for the manually curated ZJUMCH cohort set, where high-precision phenotypic descriptions aligned with symbolic clinical logic. Although the PBR engine alone exhibited suboptimal performance on the high-density PUMCH cohort, its integration into the fusion framework provided a significant complementary advantage. Notably, FusionL2 (Meta) achieved a relative improvement of more than 20\% in Recall@1 compared with the strongest single-paradigm baseline (RNC, $P<0.001$), demonstrating that neither neural pattern recognition nor symbolic reasoning alone can resolve the intrinsic complexity of RDs.

\textbf{The Multiplicative Power of Genomic Integration.}
Integrating molecular evidence served as an important diagnostic anchor (Fig.~\ref{fig:ablation}b). Transitioning from phenotype-only (FusionL2-Meta) to joint analysis (FusionL2-Joint) improved Recall@1 from 46.1\% to 76.0\%. Although both frameworks used the same underlying genotype scoring mechanism as LIRICAL (see Section~\ref{joint-method}), FusionRare achieved a higher overall diagnostic yield (Fig.~\ref{fig:gene_test}). These findings underscore that the efficacy of genomic diagnosis is not solely a function of variant analysis; rather, high-fidelity phenotypic precision acts as a mandatory catalyst that unlocks the full potential of genomic evidence.

\textbf{The Effectiveness of Agentic Reflection.}
The comparative analysis demonstrated consistent performance improvements from the knowledge-enhanced agentic reflection stage (ReflectionL3). Across all the evaluated cohorts, ReflectionL3 outperformed FusionL2 by resolving interparadigm conflicts through the synthesis of clinical descriptions with structured medical knowledge (Fig.~\ref{fig:ablation}a, b). For the Public dataset, the reflection layer improved Recall@1 from 46.4\% to 50.3\% ($P<0.01$), and for the Public-variant cohort, it increased performance from 76.0\% to 79.1\% ($P<0.001$). For the high-density PUMCH cohort, it increased Recall@1 from 45.9\% to 49.5\% ($P<0.001$). Even for the curated ZJUMCH cohort, for which improvements in Recall@1 were smaller (38.9\% to 40.1\%, $P=0.1605$),  the module refined the prioritized differential candidates. These results suggest that the agentic refinement layer functions as a decision arbiter, moving beyond statistical rank aggregation to resolve conflicting diagnostic hypotheses.

\textbf{Benefits of Foundation Model Evolution.}
Ablation of the underlying LLMs confirmed that FusionRare benefits from advancements in foundation model capabilities (Fig.~\ref{fig:ablation}c). With respect to source-level LLM ablation for FusionL2, the reasoning-focused model DeepSeek-R1 outperformed DeepSeek-V3 and GPT-4o, but lagged behind the frontier model GPT-5.5, indicating that FusionL2 remains sensitive to the reasoning capacity of the underlying language model. In the ReflectionL3 stage, we observed an ``architectural decoupling'' effect: when provided with structured CoT prompts, DeepSeek-V3 outperformed DeepSeek-R1 in the reflection task (Fig.~\ref{fig:ablation}d), suggesting that specialized prompt design can partially compensate for differences in baseline reasoning capacity. Nonetheless, GPT-5.5 achieved the highest performance for ReflectionL3 ($P<0.001$ for the improvement in Recall@1), indicating that the framework is structured to leverage future improvements in LLM capability without requiring architectural modifications.

\textbf{Cross-Scenario Robustness and Stepwise Escalation.}
The performance advantages of FusionRare were maintained across 12 medical specialties and varying levels of clinical information density (Fig.~\ref{fig:ablation}e, f). To analyse this behaviour, we tracked the diagnostic trajectory of all 2,528 cases through each stage of the pipeline (Fig.~\ref{fig:ablation}g). The statistical aggregation layer effectively synthesized consistent signals from the source engines to increase the number of correct diagnoses to the top-ranked position. The agentic refinement layer then identified and promoted correct diagnoses initially ranked 2--10 to the top-1 position. This stepwise diagnostic escalation---from raw signal capture to statistical consensus and, finally, agentic refinement---enabled the framework to resolve challenging, borderline clinical presentations.

\section{Discussion}

In AI-assisted RD diagnosis, reported performance metrics are frequently inflated by systematic evaluation biases, such as data leakage, contextual enrichment, and asymmetric knowledge advantages\cite{cheng_survey_2025,ghassemi_false_2021,zech_variable_2018}. A key contribution of this work is the establishment of a rigorous evaluation framework designed to isolate architectural innovation from these confounding factors. We addressed these potential biases through three control mechanisms: strict temporal knowledge alignment, foundation model parity, and source-level data compartmentalization. This framework ensures that the diagnostic performance of FusionRare stems primarily from multiparadigm synergy rather than the use of more recent medical databases or a more powerful backbone language model. Furthermore, representing clinical cases using structured HPO terms establishes a ``semantic decoupling'' that isolates the system from diagnostic cues---such as specialist framing and post-hoc summaries---latent in raw clinical narratives. This design ensures that the model evaluates prospective phenotypic signatures rather than exploiting linguistic shortcuts within the text. Beyond these controls, we advocate for institutional-level data isolation as a baseline standard for diagnostic AI research in RDs. Standard sample-level shuffling, where training and testing instances originate from the same institution, frequently masks data leakage. Under such validation regimes, models can internalize institutional documentation grammar or centre-specific phenotype recording preferences, which can lead to degraded generalizability when the model is evaluated based on cases from a different clinical centre. The consistent performance of FusionRare across two independent clinical cohorts (ZJUMCH and PUMCH) and a similarity-isolated public dataset indicates robust cross-institutional generalizability. For clinical deployment, this resilience, characterized by the ability to navigate diagnostic uncertainty across diverse clinical environments, is as critical as benchmark accuracy. In RD diagnosis, where the degree of phenotypic overlap is substantial and clinical descriptions are often heterogeneous, robust generalizability is a necessary prerequisite for clinical safety\cite{ghassemi_false_2021}.

The performance of FusionRare is rooted in the orthogonality of its constituent reasoning paradigms, which ensures that the inherent limitations of any single engine are mitigated by the strengths of others. Our mechanistic analysis indicates that no single paradigm is sufficient to address the full spectrum of RD diagnosis in isolation. The PBR engine leverages case-based reasoning by identifying historical clinical precedents, although its performance naturally decreases in clinical domains lacking historical analogues. Conversely, the LR engine establishes a stable symbolic baseline grounded in epidemiological priors, while the RNC classifier captures higher-order phenotypic patterns to provide broad diagnostic stability. The LLM engine complements these models by providing heuristic intuition for sparse or atypical phenotypic presentations---a capacity that scales directly with the development of underlying foundation models. Unlike autonomous agentic systems that rely on unconstrained multistep planning and evidence retrieval, which can introduce variance and cognitive drift in highly complex scenarios and are also vulnerable to data-poisoning attacks\cite{Alber2025}, FusionRare constrains heuristic LLM reasoning within a robust statistical consensus framework. By establishing a deterministic, auditable boundary through multiparadigm aggregation prior to initiating agentic reflection, the system achieves a balance between mathematical precision and heuristic flexibility.
A recent study showed that frontier general-purpose LLMs outperform specialized AI tools on medical benchmarks\cite{vishwanath_general-purpose_2026}. We validated a similar pattern within the RD diagnosis domain. FusionRare, however, overcomes this limitation through its multiparadigm fusion framework. Notably, even when restricted to a 2023 Q1 knowledge snapshot and utilizing previous-generation LLM backbones, FusionRare consistently outperformed the 2026 frontier model GPT-5.5. This suggests that the architectural integration of multiple reasoning modalities establishes a diagnostic baseline that outpaces the direct scaling of LLMs, while remaining positioned to benefit from future LLM advancements. For instance, redeploying FusionRare with updated medical knowledge and more capable LLMs further improves its performance (Fig.~\ref{fig:ablation}c, d), offering a scalable path towards shortening the ``diagnostic odyssey'' for RDs.

Beyond predictive accuracy, the clinical adoption of AI in diagnostics is often limited by the opacity of algorithmic outputs, which can foster clinician skepticism. Although FusionRare provides evidence-backed CoT reasoning and critical reflection, we address this transparency challenge by introducing an alternative diagnostic interface, RDmap. By projecting the high-dimensional phenotypic relationships of RDs onto a two-dimensional semantic manifold via dimensionality reduction, RDmap represents prioritized diagnostic candidates within a topological context. In this topological representation, spatial proximity corresponds to shared phenotypes and convergent genetic pathways, allowing clinicians to evaluate a recommended diagnosis within a broader nosological context rather than as an isolated rank in a ``black-box'' list. Preliminary qualitative evaluation with 20 clinical specialists suggested that this spatial representation facilitates diagnostic planning, as reflected by favourable usability and low cognitive load scores (measured via SUS and NASA-TLX subscales). Clinicians noted that RDmap highlighted adjacent disease clusters outside the immediate top-ranked list, prompting the consideration of differential diagnoses that might otherwise have been overlooked using traditional, nonspatial listings.

A fundamental challenge in the clinical adoption of medical AI lies in the dichotomy of trust: the risk of automation bias, where clinicians uncritically defer to algorithmic outputs, and the equally obstructive problem of algorithmic aversion, where the opacity of ``black-box'' logic breeds a profound lack of trust. This tension is particularly acute in RD diagnosis, where high baseline uncertainty makes both blind faith and undue skepticism equally problematic. Unlike rigid, opaque systems, FusionRare instantiates a dynamic human-in-the-loop framework that allows experts to designate pivotal phenotypes as ``diagnostic anchors'' and issue targeted instructions to steer the model's heuristic trajectory. Clinicians can witness how the system responds to their expert intuition. This interactivity transforms the diagnostic task from a static prediction into a collaborative cognitive exploration, empowering physicians to interrogate AI logic rather than being forced to either blindly accept or categorically reject its conclusions. Importantly, this paradigm mitigates the risk of ``AI-induced never-skilling'' among early-career clinicians by preventing the uncritical substitution of automated outputs for active reasoning\cite{ke_ai-induced_2026}. Our clinical trial revealed that FusionRare functions best as a validation partner for experienced clinicians, whose established competencies enable them to challenge and refine its recommendations---often exceeding the system’s standalone performance. By positioning FusionRare as a clinical decision support system (CDSS) subordinate to mature medical judgement, we preserve clinician accountability and foster earned trust, ensuring that the physician remains the ultimate diagnostic arbiter.

Several limitations remain to be addressed in future research. First, although the public datasets include cases from around the world, all EHR-derived data originate from clinical centres in China. Future studies should incorporate multinational, multiethnic prospective cohorts, particularly from underrepresented populations in low- and middle-income regions, to evaluate the generalizability of the framework across diverse demographics. Second, our genomic validation relied on a semisynthetic spike-in framework. While this provided a controlled testing environment to measure resilience against background variants, it did not fully replicate the complexity of raw \textit{de novo} clinical whole-exome or whole-genome sequencing data. Real-world clinical data are influenced by variable sequencing quality and a broad spectrum of variant types, presenting distinct challenges for multimodal AI diagnosis. Longitudinal validation of large-scale, prospective sequencing pipelines is an important next step. Third, the long-term impact of FusionRare on direct clinical outcomes remains to be quantified. While our preliminary usability and cognitive load assessments indicate high clinical acceptability, these metrics represent a static evaluation of diagnostic potential rather than a direct measurement of longitudinal clinical benefit. The definitive efficacy of the platform, specifically its capacity to reduce the years-long diagnostic odyssey and improve patient management outcomes, should be established through prospective, randomized controlled clinical trials. Furthermore, as diagnostic decision support systems are integrated into medical curricula, monitoring the potential for cognitive overreliance among junior clinicians is essential. Ensuring that computational tools augment, rather than erode, the independent reasoning skills of future physicians remains a critical consideration for the responsible translation of AI into clinical practice.

\clearpage
\section{Methods}\label{sec4}

\subsection{Overview}

The FusionRare ecosystem is designed to integrate automated RD case extraction, multiparadigm fusion diagnosis, and interactive clinical visualization (Extended Data Fig.~\ref{fig:system_overview}). The underlying methodology is structured into three synergistic pillars:
\textbf{An incremental case database foundation.} To overcome the limitations of static patient cohorts, we developed a multimodal, LLM-driven pipeline to extract and standardize clinical profiles from the open-access biomedical literature available via PubMed Central (PMC). These extracted cases, combined with public datasets, populate the Rare Patient Bank (RPB), which served as the empirical database for case-based reasoning and model training.
\textbf{A hierarchical multiparadigm diagnostic pipeline.} The engine processes standardized phenotypic queries represented by HPO terms using four orthogonal reasoning paradigms: (i) knowledge-driven likelihood-ratio scoring; (ii) a data-driven residual network classifier; (iii) case-driven patient retrieval; and (iv) heuristic zero-shot LLM diagnosis. The resulting diagnostic signals are mathematically synthesized via a statistical meta-aggregation layer, which is subsequently refined through an agentic reflection stage. In this final stage, an LLM uses CoT reasoning and retrieved medical evidence to resolve interengine conflicts and refine the prioritized differential diagnosis list.
\textbf{A semantic navigable disease map interface.} To align discrete algorithmic recommendations with clinical intuition, we projected the high-dimensional phenotypic relationships of RD entities onto a two-dimensional semantic manifold. Unlike traditional systems that output isolated, nonspatial candidate lists, RDmap functions as a topological reference that contextualizes the patient's phenotypic profile within the broader disease landscape. This interface enables clinicians to visually evaluate the spatial proximity between a query case and established disease clusters.

\subsection{Construction of the Unified Rare Disease Knowledge Base}\label{sec:rdkb}
To establish a rigorous foundation for diagnostic reasoning, we developed a systematic pipeline to consolidate phenotype--disease associations from OMIM and Orphanet (ORPHA) into a unified \textbf{Rare Disease Knowledge Base} (RDKB). This integration involved three sequential stages: 
(i) Phenotypic frequency harmonization: To resolve the heterogeneity of frequency annotations, we mapped qualitative HPO frequency descriptors (e.g., ``Frequent (79-30\%)'') and quantitative clinical ratios (e.g., ``3/10'') onto a continuous numerical scale $f \in [0, 1]$, enabling both probabilistic and categorical data to be processed within a unified statistical framework. (ii) Cross-database mapping: We leveraged direct cross-references provided by Orphanet and Mondo to map equivalent ORPHA and OMIM codes to a unified RD entity, thereby resolving discrepancies between distinct clinical coding systems. Ambiguous or conflicting mappings were manually curated and resolved to ensure high-confidence disease alignment. (iii) Knowledge consolidation and enrichment: we merged disease-phenotype annotations using a union and maximum-frequency strategy. For phenotype-disease pairs represented in both source databases, we retained the maximum frequency value to capture a broadest clinical spectrum and maximize diagnostic sensitivity. 

\subsection{Construction of the Rare Disease Patient Bank}
To address the gap between static ontological knowledge and the high-dimensional heterogeneity of real-world clinical presentations, we constructed the \textbf{Rare Patient Bank} (RPB). This repository aggregates 38,729 validated RD cases (detailed in Extended Data Table~\ref{tab:case_sources_final}), serving as the reference empirical database for case-based reasoning and model training.

\subsubsection{Automated Case Extraction and Enrichment}
The RPB is incrementally expanded through an LLM-based pipeline designed to extract RD cases from PMC (see an example in Appendix~A). While conceptually building upon the RareArena framework\cite{chen_rarearena_2026}, our pipeline implements three core architectural optimizations:

\begin{enumerate}
\item High-Throughput Pre-filtering: Articles are screened at the title and abstract levels on the basis of RDKB disease entities to minimize computational overhead and filter for high-relevance clinical narratives.
\item Multimodal Deep Parsing: The pipeline performs deep structural parsing of full-text content, specifically extracting patient profiles from complex structured tables and aligning in-text clinical figures with corresponding narrative descriptions.
\item Traceable Cross-Referencing: Each extracted clinical profile is programmatically linked to its source PMID, ensuring end-to-end traceability and clinical auditability.
\end{enumerate}

\subsubsection{Knowledge-Augmented Phenotypic Standardization}
To resolve the semantic gap between unstructured clinical notes and structured HPO terms, we implemented a multistage standardization protocol that integrates dictionary- and transformer-based\cite{vaswani_attention_2017} concept recognition with deep semantic embeddings. The workflow begins with a hybrid preprocessing phase, in which raw clinical text (in English or Chinese) is parsed using FastHPOCR\cite{groza_fasthpocr_2024} or PBTagger\cite{mao_phenotype-based_2025}. The resulting candidate entities are injected into an LLM context to serve as referential anchors. By utilizing CoT prompting (see $\langle\text{prompt}\rangle_{1}$ in Appendix~E), the LLM synthesizes these anchors into a structured summary of the patient's clinical abnormalities, ensuring that the extracted phenotypes preserve their original clinical context. Finally, HPO mapping is achieved via semantic entity resolution. We utilize the BioLORD model\cite{remy_biolord-2023_2024}---pretrained with large-scale biomedical and clinical language representations---to generate normalized embeddings for each extracted phenotypic description. These embeddings are compared against the entire HPO corpus using cosine similarity. To ensure mapping fidelity, a candidate concept is only mapped to an HPO term if the cosine similarity exceeds a threshold of 0.8. This hybrid approach combines the contextual synthesis of generative models with the mathematical precision of high-dimensional embedding spaces.

\subsection{Multiparadigm Fusion Diagnosis Schema} \label{sec:rank}
Problem Definition: Let $\mathcal{Q} = \{ t_1, t_2, ..., t_{n} \}$ be the set of HPO terms observed in a patient. The goal of the schema is to produce a prioritized differential diagnosis list $\mathcal{L}=\{(d_i, \text{score}_i) \mid d_i \in \mathcal{D} \}$, where disease $d_i$ belongs to the disease space $\mathcal{D}$ defined in RDKB. Our approach follows a four-stage pipeline: (i) multisource phenotypic evidence generation, (ii) statistical phenotypic rank aggregation, (iii) optional phenotype-genotype joint reranking, and (iv) knowledge-grounded CoT reflection.

\subsubsection{Source-level Diagnosis Approaches}
We first employ four heterogeneous diagnostic engines to capture different facets of phenotypic evidence.

\textbf{(1). Likelihood Ratio (LR) Scoring (Knowledge-Driven):} The Bayesian evidence for each candidate disease $d$ given $\mathcal{Q}$ is calculated based on RDKB in this method. First, the probability of the HPO term $t$ for disease $d$ is calculated as follows:
\begin{equation}
\Pr(t \mid d) = \begin{cases} 
f^t_d & \text{if } t \in \Anno(d) \\ 
1 - \prod_{t' \in \Anno(d)\cap \Desc(t)} [1 - f^{t'}_d], & \text{if } t \in \Anc(\Anno(d)) \setminus \Anno(d) \\ 
0, & \text{otherwise} 
\end{cases}
\end{equation}
where $f^t_d$ is the numerical frequency of phenotype $t$ for disease $d$ in RDKB, $\Anno(d)$ represents the annotated HPOs of $d$, and $\Anc(\cdot)$ and $\Desc(\cdot)$ represents the inclusive ancestor and descendant HPO sets, respectively. Assuming conditional independence of phenotypes given the disease, the LR score of query $\mathcal{Q}$ for disease $d$ is:
\begin{equation}
\LR(\mathcal{Q}, d) = \prod_{t_i \in Q} \LR(t_i, d) = \prod_{t_i \in Q} \frac{\Pr(t_i\mid d)}{\Pr(t_i\mid\bar{d})}
\end{equation}
where $\Pr(t_i\mid\bar{d})$ is approximated by the background frequency $\sum_{d \in \mathcal{D}} \Pr(t_i\mid d) / |\mathcal{D}|$, representing the expected prevalence of $t_i$ across the disease space. By Bayes' theorem, the posterior odds of disease $d$ given query $\mathcal{Q}$ is:
\begin{equation}
\Odds_\text{posterior}(d \mid \mathcal{Q}) = \Odds_\text{prior}(d) \cdot\LR(\mathcal{Q}, d)
\end{equation}
Treating $\Odds_\text{prior}(d)$ as a constant across all candidate diseases, the log-posterior odds is proportional to the sum of the individual log-likelihood ratios $\log\LR(t_i, d)$:
\begin{equation}
\log \Odds_\text{posterior}(d \mid \mathcal{Q}) \propto \sum_{t_i \in \mathcal{Q}}\log\LR(t_i,d)
\end{equation}
This computation yields the ranked list $\mathcal{L}_\text{LR}$, prioritizing diseases by their cumulative LR scores.

\textbf{(2). Residual Network Classifier (RNC) (Data-Driven):} The RNC bridges explicit medical knowledge and implicit clinical patterns by integrating a fixed knowledge-based prior branch with a learnable residual branch. We define a multi-hot query vector $\mathbf{x} \in \{0, 1\}^{M\times 1}$ indicating the presence of query HPO terms among all $M$ identifiable HPOs, where $\mathbf{x}_i=1$ if corresponding HPO term $t_i \in \mathcal{Q}$. 

\textbf{Prior Knowledge Branch.} This branch utilizes a fixed weight matrix $\mathbf{W}_\text{LR} \in \mathbb{R}^{N \times M}$ ($N=|\mathcal{D}|$), where $w_{ij}=\log \LR(t_j, d_i)$. The linear transformation $ \mathbf{W}_\text{LR}\mathbf{x}$  effectively performs batch Bayesian inference in a single pass.

\textbf{Residual Architecture.} To capture nonlinear phenotypic correlations beyond RDKB, we implement the residual branch $\Delta \mathbf{z}$ as a Multi-Layer Perceptron (MLP) with a hidden dimension $H=1,000$:
\begin{equation}
 \mathbf{z} = \underbrace{\mathbf{W}_\text{LR}\mathbf{x}}_{\text{Prior Log-Odds}} + \alpha \cdot \underbrace{\left( \mathbf{W}_2(\sigma(\mathbf{W}_1 \tilde{\mathbf{x}} + \mathbf{b}_1)) + \mathbf{b}_2 \right)}_{\text{Learned Residual } \Delta \mathbf{z}}
\end{equation}
where $\{\mathbf{W}_1 \in \mathbb{R}^{H \times M}, \mathbf{b}_1 \in \mathbb{R}^{H\times 1}\}$ and $\{\mathbf{W}_2 \in \mathbb{R}^{N \times H}, \mathbf{b}_2 \in \mathbb{R}^{N\times 1}\}$ denote the learnable weights and biases of the hidden and output layers, respectively. $\sigma(\cdot)$ represents the ReLU activation. The learnable scalar $\alpha$ modulates the trade-off between the knowledge-based prior and the data-driven correction. The parameter set is $\Theta=\{\mathbf{W}_1, \mathbf{b}_1, \mathbf{W}_2, \mathbf{b}_2, \alpha\}$.
$\tilde{\mathbf{x}}$ is the ancestral expansion of $\mathbf{x}$, where $\tilde{\mathbf{x}}_j=1$ if the corresponding HPO term $t_j \in \{t_i\cup \Anc(t_i)\mid t_i\in \mathcal{Q}\}$. The output $\mathbf{z} \in \mathbb{R}^{N\times 1}$ provides the final logits for all diseases. This structure ensures that the model utilizes the LR-based prior as a baseline while the deep learning component refines predictions based on cohort patterns.

\textbf{Training Strategy.} The RNC model is optimized using the Cross-Entropy loss function and the AdamW optimizer with a learning rate of 1e-5. Most cases from RPB were included in the training set. Considering that RPB does not cover all the diseases in RDKB, we augmented the empirical training data by generating 10 synthetic cases for each RD per training epoch (see details in Appendix~B). This strategy enables the classifier to learn generalized diagnostic patterns across the entire disease space, while maintaining high precision on frequently encountered clinical cases.

The RNC outputs a ranked list $\mathcal{L}_\text{RNC}$ prioritized by these integrated disease logits.

\textbf{(3). Patient Bank Retriever (PBR) (Case-Driven):} The PBR retrieves diagnoses by measuring the phenotypic similarity between the query $\mathcal{Q}$ and historical cases $\mathcal{Q}_\text{ref}$ within the case space $\mathcal{P}$ of the RPB. The information content (IC) of an HPO term $t$, representing its diagnostic specificity, is calculated as follows:
\begin{equation}
\IC(t) = -\log \left( \frac{\sum_{d \in \mathcal{D}} \Pr(t \mid d)}{|\mathcal{D}|} \right)
\end{equation}
The phenotypic similarity between $\mathcal{Q}$ and a historical case $\mathcal{Q}_\text{ref}$ with a known diagnosis is calculated as the average shared IC:
\begin{equation}
\Sim(\mathcal{Q}, \mathcal{Q}_\text{ref}) = \frac{1}{2} \left[ \sum_{t \in \mathcal{Q} \cap \Anc(\mathcal{Q}_\text{ref})} \IC(t) + \sum_{t' \in \mathcal{Q}_\text{ref} \cap \Anc(\mathcal{Q})} \IC(t') \right]
\end{equation}
To derive a disease-level ranking from case-level similarities, we apply a max-pooling strategy. Let $\mathcal{C}_d$ be the set of historical cases in RPB diagnosed with disease $d$. The priority score for disease $d$ is defined by its most similar historical presentation:
\begin{equation}
\Sim_\text{PBR}(\mathcal{Q}, d) = \max_{\mathcal{Q}_\text{ref} \in \mathcal{C}_d} \Sim(\mathcal{Q}, \mathcal{Q}_\text{ref})
\end{equation}

This yields the ranked list $\mathcal{L}_\text{PBR}$, which prioritizes diseases if the query patient closely resembles at least one documented clinical instance.

\textbf{(4). Zero-Shot LLM Diagnosis (Heuristic-Driven):} This approach leverages the parametric medical knowledge of an LLM for heuristic inference. Given $\mathcal{Q}$, the LLM is prompted to generate a differential diagnosis list $\mathcal{L}_\text{LLM}$ containing the top 10 candidate diseases (see details of the prompt in Appendix~E):
\begin{equation}
\mathcal{L}_\text{LLM} = f_\text{LLM}(\mathcal{Q};\langle\text{prompt}\rangle_{2})
\end{equation}

To integrate this output quantitatively, each disease in $\mathcal{L}_\text{LLM}$ is assigned a heuristic, linearly decreasing confidence score ranging from $1.0$ (top‑ranked) to $0.5$ (the lowest).
To resolve the nonstandard nomenclature in the LLM-generated diagnoses, we map each candidate disease to a canonical identifier using semantic entity resolution. Specifically, we embed both the LLM-generated diseases and all RDKB disease names with BioLORD, then assign the canonical identifier of the highest cosine similarity match.

Let $\mathcal{L}'=\{\mathcal{L}_\text{LR}, \mathcal{L}_\text{RNC}, \mathcal{L}_\text{PBR}, \mathcal{L}_\text{LLM}\}$ denote the set of ranked lists generated by these four source-level approaches.

\subsubsection{Phenotypic Diagnosis Ranking Aggregation}
To synthesize the heterogeneous outputs from source-level approaches $\mathcal{M}=\{\text{LR},\text{RNC},\text{PBR},\text{LLM}\}$, we propose a two-stage hierarchical fusion framework. Let $\mathcal{L}_m$ be the ranked diagnosis list generated by method $m \in \mathcal{M}$, with $r_m(d)$ and $s_m(d)$ denoting the 1-based rank and raw priority score of disease $d$, respectively.
For each candidate diagnosis $d$, we define its normalized rank score $x_m(d)$ as follows:
\begin{equation}
x_m(d) = \frac{r_m(d)}{N_m}
\end{equation}
where $N_m$ is the total number of candidate diagnoses in $\mathcal{L}_{m}$. If $d \notin \mathcal{L}_{m}$, it is penalized with $x_m(d) = 1.0$.

\textbf{(1). Base Aggregation:}
We calculate intermediate consensus scores for each candidate using four distinct aggregation strategies:

\begin{itemize}

\item \textbf{Minimum Rank\cite{kolde_robust_2012}:} Captures the best individual model performance to maximize expected recall:
\begin{equation}
S_\text{Min}(d) = \min_{m \in \mathcal{M}} x_m(d)
\end{equation}

\item \textbf{Geometric Mean Rank\cite{kolde_robust_2012}:} Favors candidates with consistent performance across multiple models while dampening outliers:
\begin{equation}
    S_\text{Geom}(d) = \exp \left( \frac{1}{|\mathcal{M}|} \sum_{m \in \mathcal{M}} \log x_m(d) \right)
\end{equation}

\item \textbf{Distribution-based Rank\cite{kolde_robust_2012}:} Assuming $x_{m}$ follows a uniform distribution $U(0,1)$ under the null hypothesis, the average rank $\bar{x}(d) = \frac{1}{|\mathcal{M}|} \sum_{m \in \mathcal{M}} x_{m}(d)$ is modelled using a normal approximation of the Bates distribution:
\begin{equation}
S_\text{Bates}(d) = \Phi(\frac{\bar{x}(d) - 0.5}{\sqrt{1 / (12 \cdot |\mathcal{M}|)}})
\end{equation}
where $\Phi(\cdot)$ is the cumulative distribution function of the standard normal distribution.

\item \textbf{Score Evidence-based Rank\cite{renda_web_2003}:} Unlike normalized rank score-based methods, this strategy uses the raw priority score $s_m(d)$. Let $z_m(d) = (s_m(d) - \mu_m) / \sigma_m$ be the standard score (Z-score) of $s_m(d)$ within $\mathcal{L}_m$. To reward diseases backed by higher absolute evidence magnitude and broader cross-method support, the integrated score is defined as follows:
\begin{equation}
S_\text{CombMNZ}(d) = |\{m : d \in \mathcal{L}_m\}| \cdot \sum_{m \in \mathcal{M}} z_m(d)
\end{equation}
\end{itemize}

For prioritization, candidate lists generated by $S_\text{Min}$, $S_\text{Geom}$ and $S_\text{Bates}$ are sorted in ascending order, while $S_\text{CombMNZ}$ is in descending order.

\textbf{(2). Meta-aggregation:}
To reduce the methodological variance inherent to any single aggregation strategy, we perform a secondary meta-aggregation across the base aggregators $\mathcal{M'}=\{\text{Min}, \text{Geom}, \text{Bates}, \text{CombMNZ}\}$.
We integrate their outputs via a weighted geometric mean. Let $x_{m'}(d)$ denote the new normalized rank score of $d$ within the list generated by strategy $m' \in \mathcal{M'}$. The final meta-score $S_\text{Meta}(d)$ is formulated as follows:
\begin{equation}
S_\text{Meta}(d) = \exp \left( \sum_{m' \in \mathcal{M'}} w_{m'} \log x_{m'}(d) \right)
\end{equation}
where $w_{m'} \in [0, 1]$ is the weight assigned to aggregator $m'$. We utilized the Optuna framework to optimize the continuous weight distribution $\mathbf{w} \in [0, 1]^{|\mathcal{M'}|}$ on an independent validation set, maximizing the cumulative diagnostic recall ($\text{Recall@5}$ and $\text{Recall@10}$).
To further enhance the model's generalizability and mitigate the risk of overfitting to specific weight fluctuations, continuous weights were subsequently quantized to a discrete set $\{1.0, 0.5, 0\}$ based on their nearest proximity. For details, see Appendix~C.

This meta-aggregation approach provides a smoothing effect, reducing sensitivity to outliers in both the source models and the first-level aggregators. The resulting prioritized list $\mathcal{L}''$, sorted in ascending order by $S_\text{Meta}$, represents the statistical consensus of the multiparadigm fusion schema.

\subsubsection{Phenotype-Genotype Joint Diagnosis}\label{joint-method}
When genomic sequencing data is available, we incorporate molecular evidence to refine the phenotypic consensus $\mathcal{L}''$. The LIRICAL framework computes an LR score for each candidate pathogenic genotype $\mathcal{G}=\{g_1, g_2, ..., g_n\}$ from the input VCF, and assigns the score to the corresponding disease based on disease-genotype associations. The joint diagnostic score $S_\text{Joint}(d)$ integrates the log-transformed genotype evidence with the phenotypic evidence $S_\text{Meta}$. To prevent uninformative or negative genotype evidence from penalizing strong phenotypic consensus, we apply minimum-value clipping to the genotype evidence:
\begin{equation}
S_\text{Joint}(d) = -\log S_\text{Meta}(d) + \max\bigl(\log\LR^*_\text{geno}(d), 0\bigl)
\end{equation}
where $\LR^*_\text{geno}(d)$ is the genotype LR score for disease $d$. Candidate diseases are then reranked in descending order of $S_\text{Joint}$, yielding a genotype-adjusted consensus list $\mathcal{L}''$.

\subsubsection{Knowledge-Enhanced CoT Reflection}
The final stage involves a knowledge-grounded CoT reflection. As shown in Eq. (\ref{eq:reflection}), the LLM acts as a clinical expert to synthesize source-level evidence $\mathcal{L}'$, the statistical consensus $\mathcal{L}''$, and the structured medical knowledge 
$\mathcal{K}$ from RDKB. By reasoning through conflicts between different diagnostic paradigms, the LLM produces a refined top-5 candidate list $\mathcal{L}^*_5$ (see details of the prompts in Appendix~E).
\begin{equation}
\mathcal{L}^*_5 = f_\text{LLM}(\mathcal{Q}, \mathcal{L}', \mathcal{L}'', \mathcal{K};\langle\text{prompt}\rangle_{3}, \langle\text{prompt}\rangle_{4})
\label{eq:reflection}
\end{equation}

To ensure the final differential diagnosis list $\mathcal{L}'''$ maintains high diagnostic coverage, we concatenate the LLM-optimized candidates $\mathcal{L}^*_5$ with the remaining diseases from the statistical consensus (Eq.~\ref{eq:final_rerank}). Here, $\parallel$ denotes sequence concatenation, and $\ominus$ represents the set difference operation while preserving the relative order of the remaining elements.
\begin{equation}
\mathcal{L}''' = \mathcal{L}^*_5  \parallel (\mathcal{L}'' \ominus \mathcal{L}^*_5)
\label{eq:final_rerank}
\end{equation}

\subsection{Map Interface Derived from RDKB}
To support high-dimensional data retrieval and diagnostic exploration, we developed a map-based semantic interface (RDmap). This interface projects discrete, high-dimensional phenotypic annotations of diseases into a continuous topological space via the construction of a global similarity-based manifold.

\textbf{(1). Global Similarity Matrix Construction:}
Let $\mathcal{D}=\{d_i\}^N_{i=1}$ denote the set of disease entities in RDKB. We construct a pairwise similarity matrix $\mathbf{S} \in \mathbb{R}^{N \times N}$, where each element $s_{ij}$ represents the phenotypic proximity between disease $d_i$ and $d_j$. The $i$-th row vector $\mathbf{s}_i=[s_{i1}, s_{i2}, ..., s_{iN}]$ effectively serves as a high-dimensional relational embedding of disease $d_i$, capturing its relative position within the global RD universe.

\textbf{(2). Similarity Engine (Lin-BMA):}
The elements $s_{ij}$ of the matrix are computed via an ontology-grounded semantic engine. Let $\mathcal{Q}_i$ and $\mathcal{Q}_j$ represent the related HPOs of $d_i$ and $d_j$ respectively, then $s_{ij}$ is calculated using the Best Match Average (BMA) of term-level Lin similarity\cite{lin_information-theoretic_1998}:
\begin{equation}
s_{ij} = \frac{1}{2} \left[ \frac{1}{|\mathcal{Q}_i|} \sum_{t_u \in \mathcal{Q}_i} \max_{t_v \in \mathcal{Q}_j} \Sim_\text{Lin}(t_u, t_v) + \frac{1}{|\mathcal{Q}_j|} \sum_{t_v \in \mathcal{Q}_j} \max_{t_u \in \mathcal{Q}_i} \Sim_\text{Lin}(t_u, t_v) \right]
\end{equation}
where $\Sim_\text{Lin}(t_u, t_v)= 2\cdot \IC(\MICA(t_u, t_v))/(\IC(t_u)+\IC(t_v))$, and $\MICA(t_u, t_v)$ is the Most Informative Common Ancestor of HPO terms $t_u$ and $t_v$.

\textbf{(3). Manifold Learning and Visualization:}
To characterize the latent manifold geometry inherent in the relational similarity vectors $\{\mathbf{s}_i\}^N_{i=1}$, we utilize Uniform Manifold Approximation and Projection (UMAP)\cite{mcinnes_umap_2018} to compute a nonlinear embedding from the $N$-dimensional relational space to a two-dimensional Euclidean space $\mathbb{R}^2$. The resulting projection maps the disease space $\mathcal{D}$ to a two-dimensional coordinate system, in which the spatial proximity of disease nodes reflects their underlying phenotypic similarity. To optimize the clinical interpretability of the manifold, each node is annotated with multidimensional metadata: (i) Chromatic encoding: Disease nodes are coloured either statically on the basis of established medical taxonomies or dynamically to highlight the prioritized candidates generated for a specific query; (ii) Geometric scaling: The visual radius of each disease node is scaled proportionally to its epidemiological prevalence.

\subsection{Clinical Diagnostic Study Design}
We conducted a blinded, two-stage evaluation to quantify the incremental diagnostic value of FusionRare over current clinical practice (Fig.~\ref{fig:trial}a). Participating physicians from multiple specialties---including genetic metabolism, endocrinology, cardiology, rheumatology, and neonatology---were recruited over a one-month period. The final cohort comprised 20 clinicians (mean age 41 years; range 29–58), randomly assigned to two balanced groups (Clinician Group A/B) to control for baseline differences in diagnostic expertise. Fifty RD cases, drawn from the established challenging subset of our benchmark, were used as the diagnostic test set. Each case represented a distinct disease to maximize phenotypic coverage and ensure no diagnostic overlap. The cases were randomly divided into two equal subsets (Case Set A/B), with diagnostic difficulty balanced across groups.

\textbf{(1). Routine clinical evaluation (Stage I): }
In the first stage, physicians independently evaluated their assigned case set without access to FusionRare, relying solely on conventional clinical resources. These included medical textbooks, internet search engines, knowledge bases, and publicly available LLMs (e.g., DeepSeek, ChatGPT, Doubao). For each case, clinicians were required to provide up to three ranked differential diagnoses.

\textbf{(2). FusionRare-assisted evaluation (Stage II): }
In the second stage, all participants received a standardized introduction to FusionRare and its workflow. This stage comprised two sequential sub-assessments:
\begin{enumerate}
\item Re-evaluation: Participants re-assessed the same cases from Stage I using FusionRare alongside their routine resources, again providing up to three reranked diagnoses. For each case, participants additionally reported changes in subjective diagnostic confidence relative to their initial assessment, using a five-point ordinal scale: (i) persistently low confidence, (ii) increased from low to moderate confidence, (iii) persistently moderate confidence, (iv) increased from moderate to high confidence, and (v) persistently high confidence. This measure served as a proxy for perceived diagnostic certainty and clinical trust in FusionRare.
\item Crossover evaluation: Participants evaluated the alternative, previously unseen case set with FusionRare.
\end{enumerate}

Throughout both stages and all sub-assessments, strict blinding was maintained: the final diagnoses were withheld from participants, and no performance feedback was provided until study completion. This design allowed us to isolate the effect of FusionRare from learning effects or case familiarity.
After completing both stages of the study, participants completed two validated instruments to assess clinical applicability. Cognitive workload was measured using the NASA Task Load Index (NASA-TLX), capturing mental demand, temporal demand, effort, and frustration. System usability was assessed using the System Usability Scale (SUS), a standardized 10-item questionnaire yielding a composite usability score. The detail evaluation forms were shown in Appendix~D. These scales provided complementary insights into clinician acceptance, perceived workload, and overall user experience. All participant responses were manually verified and curated by the research team.

\subsection{Benchmarking and Statistical Settings}

\textbf{Modalities of implementation.} To ensure reproducibility and version control, baseline methods were deployed across three modalities: (i) local containerization, where standalone implementations of Exomiser, LIRICAL, PhenoPro, BOQA, and PhenoDP were executed via official source code or binary distributions; (ii) API facilitation, where PhenoBrain and generative LLMs were accessed via developer API endpoints; and (iii) web-interface execution, where DeepRare was evaluated through its official portal (https://raredx.cn/).

\textbf{Experimental fairness and version control.} To maintain identical knowledge constraints acrossthe models, including FusionRare, all the controllable baselines were synchronized to the 2023 Q1 release of OMIM/Orphanet and aligned with the knowledge snapshot of the PhenoBrain baseline. For the underlying foundation models, FusionRare utilized DeepSeek-R1 and DeepSeek-V3, which are consistent with DeepRare's backbone LLMs.

\begin{figure}[ht!]
    \captionsetup{name=Extended Data Fig.}
    \centering
    \includegraphics[width=1.0\textwidth]{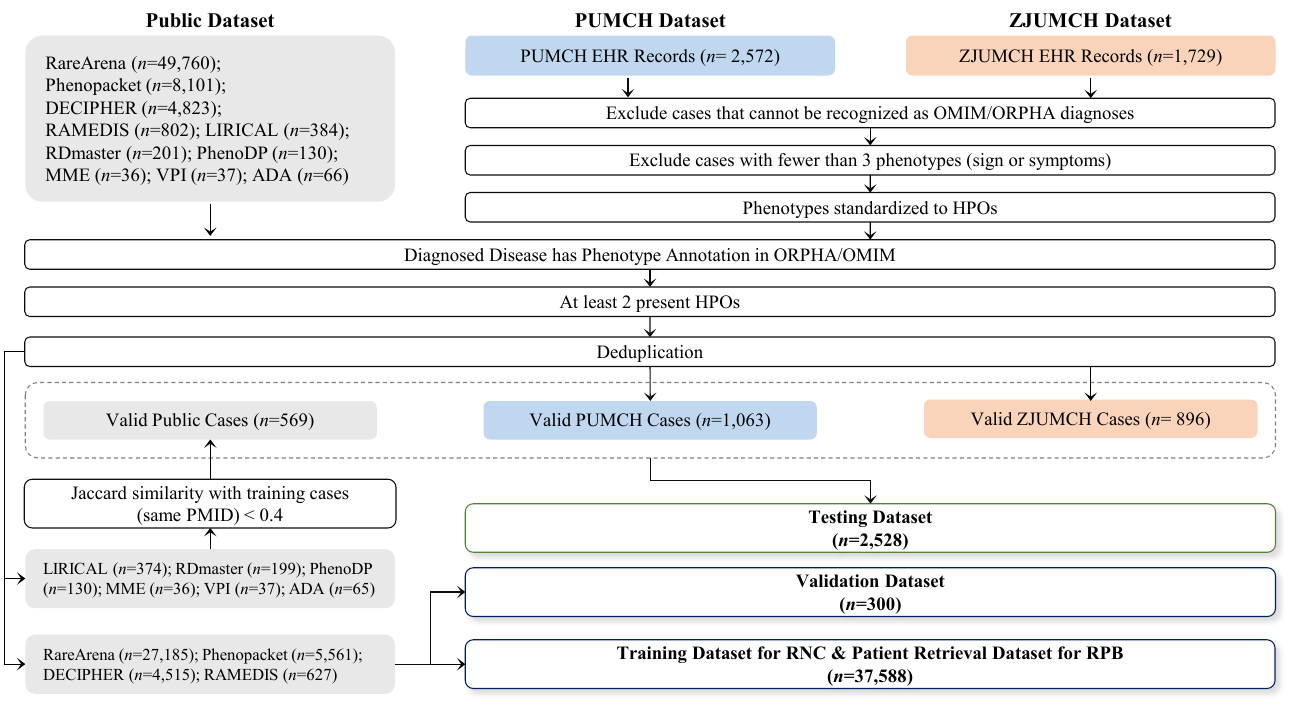} 
    \captionof{figure}{\textbf{Details of RD case dataset curation.} The flowchart delineates the systematic pipeline for data standardization, quality control, and rigorous isolation between the developmental and evaluation sets. \textbf{(i)} Data acquisition and preliminary filtering. Initial patient records ($n=63,417$ across 12 sources) were aggregated from public literature, web registries, and multicentre EHR systems (PUMCH and ZJUMCH). To ensure diagnostic fidelity, cases were excluded if their clinical diagnoses were not mappable to the RDKB (OMIM/ORPHA). \textbf{(ii)} Quality control and standardization. Phenotypic profiles were subjected to a minimum-density threshold, requiring at least three clinical descriptors (signs or symptoms). All free-text clinical notes were standardized to HPO terms. To align with the scope of the diagnostic engine, we excluded cases with ground-truth diseases that lacked phenotype annotations in the RDKB. \textbf{(iii)} Anti-leakage and strategic partitioning. After global deduplication, the datasets were stratified on the basis of provenance and utility. The training/retrieval dataset ($n=37,588$) was exclusively sourced from the literature and web registries (e.g., RareArena and DECIPHER) to populate the RNC classifier and RPB index. For the Public Test Cohort ($n=569$), a secondary similarity-based filter was applied: any case sharing a PMID with the training set was retained only if its HPO Jaccard similarity to all training instances was $< 0.4$, effectively precluding pattern memorization. \textbf{(iv)} Independent clinical validation. The Testing Dataset ($n=2,528$) comprised the filtered Public cases and two strictly isolated EHR cohorts (PUMCH and ZJUMCH). Within this testing dataset, a Challenge Subset ($n=300$) was randomly sampled for clinician-led manual audits (100 cases per source). This multilayered isolation ensured that the evaluation measured genuine cross-institutional generalizability rather than data-driven artefacts.}
    \label{fig:dataset}
\end{figure}

\textbf{Data compartmentalization for PBR.} For case-based reasoning with the PBR engine, we enforced strict data isolation. Evaluations on the Public test set utilized strictly source-isolated reference cases (Extended Data Fig.~\ref{fig:dataset}), whereas in evaluations using clinical cohort data (ZJUMCH and PUMCH), the full public dataset was used as the reference retrieval bank.

\textbf{Harmonization of genetic diagnosis scale.} Because diagnostic outputs vary between gene-centric (e.g., Exomiser) and disease-centric (e.g., LIRICAL, FusionRare) models, we implemented a bidirectional conversion protocol to ensure a fair comparison. For gene-centric models, gene-level ranks were expanded into disease-level ranks by mapping each candidate gene to its associated OMIM and ORPHA terms. Conversely, for disease-centric models, disease rankings were mapped to gene-level ranks by substituting predicted diseases with their causative genetic variants identified in sequencing data, followed by stable deduplication. This ensured that both gene and disease prioritizations were evaluated on an identical scale.

\textbf{Metrics and performance analysis.} 
The ground-truth diagnoses for all test cases were defined by OMIM or ORPHA codes. Benchmarks were evaluated using an automated code-matching paradigm except the clinical consistency evaluation on the challenge subset and the clinical trial. Diagnostic efficacy was quantified using Recall@$n$ ($n \in [1, 10]$). For cases with multiple causative diagnoses, we employed an any-hit strategy, where a test was recorded as a success if any true causative entity was present within the top-$n$ recommendations. We incorporated the cross-database mappings described in Section~\ref{sec:rdkb} to ensure comparability: (i) a predicted ORPHA code was counted as a match if it mapped to the true-diagnosed OMIM entry, and vice versa; and (ii) for models restricted to a single nomenclature (ORPHA or OMIM), we applied cross-database mapping to deduplicate their candidate lists. Crucially, to prevent evaluation bias, models with restricted outputs (e.g., DeepRare, which outputs only the top-5 candidates) were penalized; if the ground-truth diagnosis was absent from the provided candidate list or was ranked outside the top 10, the case was classified as a miss.

\textbf{Statistical Analysis.} 
For diagnostic performance evaluations and ablation analyses, we performed a nonparametric bootstrap resampling method with 2,000 iterations\cite{efron_introduction_1994}. For each iteration, $n$ cases were resampled with replacement from the respective benchmark ($n=$size of the respective benchmark). The 95\% confidence interval (CI) was derived from the 2.5th and 97.5th percentiles of the bootstrap distribution. $P$-values were calculated as the proportion of bootstrap replicates where a benchmarking model's Recall@{k} matched or exceeded that of the baseline model (typically FusionRare, one-sided test), following the methodology described by Koehn\cite{koehn_statistical_2004}.
For the clinical trial evaluation, statistical comparisons were conducted based on the individual diagnostic accuracies (Recall@$k$) of the participating clinicians. Specifically, the Wilcoxon signed-rank test (for paired comparisons, same clinician with vs. without FusionRare) and the Mann-Whitney U test (for independent comparisons, routine evaluation vs. crossover evaluation) were performed using the \texttt{scipy.stats} module in Python to compute $P$-values. Effect sizes were quantified using Cohen's $d_z$ for paired comparisons\cite{lakens_calculating_2013}.

\backmatter

\section*{Data Availability}
The Human Phenotype Ontology (HPO) and its disease annotations are publicly available for download from the official HPO portal (https://hpo.jax.org/). Similarly, disease annotation data from Orphanet (ORPHA; https://www.orpha.net/) and the Mondo Disease Ontology (Mondo; https://mondo.monarchinitiative.org/) can be accessed via their respective official portals. For the clinical datasets used in this study, the download links and original references for all publicly accessible cohorts (including PUMCH) are provided in Extended Data Table~\ref{tab:data_sources}. To support reproducibility, the 300 deidentified case profiles from the challenge subset, along with their benchmark performance data, are publicly available on Zenodo at https://zenodo.org/records/20697776. The primary electronic health record (EHR) data from the Children's Hospital of Zhejiang University School of Medicine (ZJUMCH) are not publicly available because of institutional data privacy regulations and ethical constraints. All ZJUMCH patient data utilized in this study were processed under institutional review board (IRB) approval (ZJUMCH IRB \#2021-IRB-165, \#2022-IRB-068) and were fully deidentified prior to computational analysis.

\section*{Code Availability}

The source code is available at https://github.com/1678492347/FusionRare. The extended web application developed for this research is hosted at http://rdx.nbscn.org/.

\section*{Acknowledgements}

This work is supported by the Natural Science Foundation of China (Nos. 62506329 and 82572377) and the ``Pioneer'' and ``Leading Goose'' R\&D Program of Zhejiang (No. 2026C01022).

\section*{Author Contributions}

All listed authors meet the four ICMJE criteria for authorship.
All the authors (J.Y., P.L., S.L., Y.C., Y.F., L.Y., J.P., J.Ye, X.H., C.Z., X.L., H.D., and H.L.) contributed to the design and conduct of the study.
J.Y. and P.L. led the algorithm prototype design.
H.L., C.Z., X.H., J.Ye, and J.P. led the clinical design and medical aspects of the study.
J.Y., P.L., L.Y., and H.L. performed data acquisition, analysis, and interpretation.
P.L., J.Y, S.L., and H.L. contributed to the implementation of the web platform.
J.Y. and H.L. contributed to the evaluation framework used in the study.
H.L., H.D., and X.L. provided technical and infrastructure guidance.
J.Y., P.L., and H.L. drafted the manuscript, and H.L., S.L., Y.C., and Y.F. provided critical revisions for important intellectual content.
All the authors approved the final version of the manuscript for publication and agree to be accountable for all aspects of the work and ensure that questions related to the accuracy or integrity of any part of the work are appropriately investigated and resolved.

\section*{Competing Interests}

We declare that the authors have no competing interests as defined by Nature Portfolio, or other interests that might be perceived to influence the results and/or discussion reported in this paper.

\clearpage
\bibliography{fusionrare}

\clearpage
\begin{figure}[ht!]
    \captionsetup{name=Extended Data Fig.}
    \setcounter{figure}{0}
    \centering
    \includegraphics[width=1.0\textwidth]{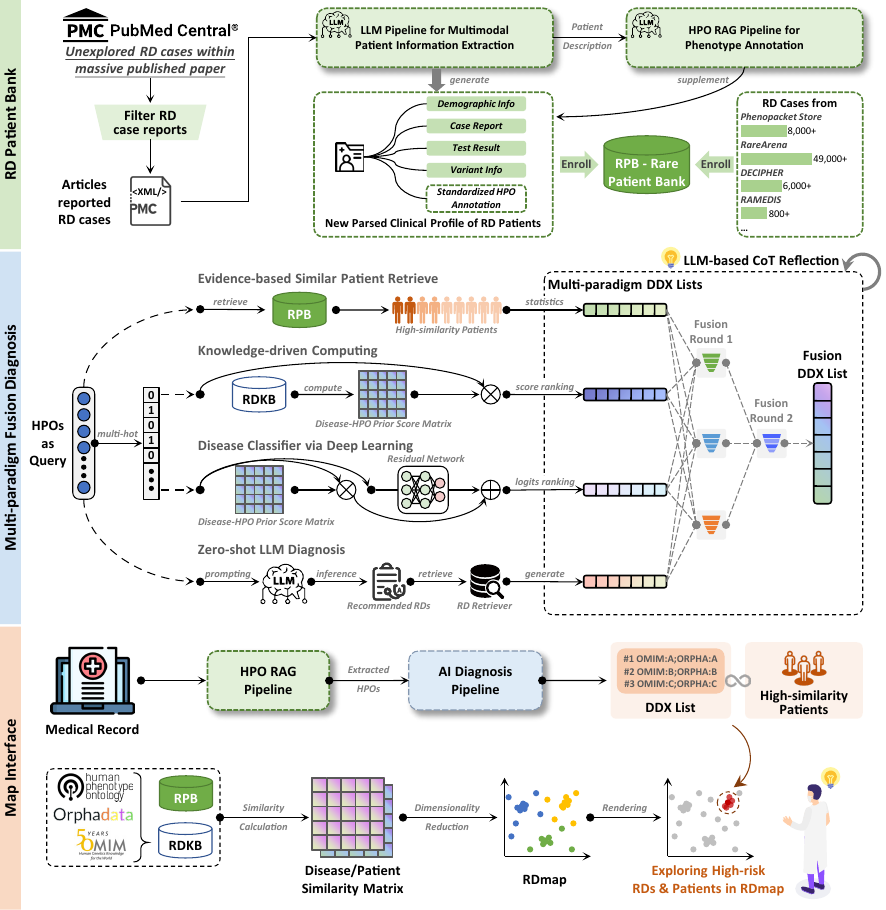} 
    \caption{\textbf{Overview of FusionRare system.}}
    \label{fig:system_overview}
\end{figure}

\clearpage

\clearpage
\begin{figure}[h!]
    \captionsetup{name=Extended Data Fig.}
    \centering
    \includegraphics[width=1.0\textwidth]{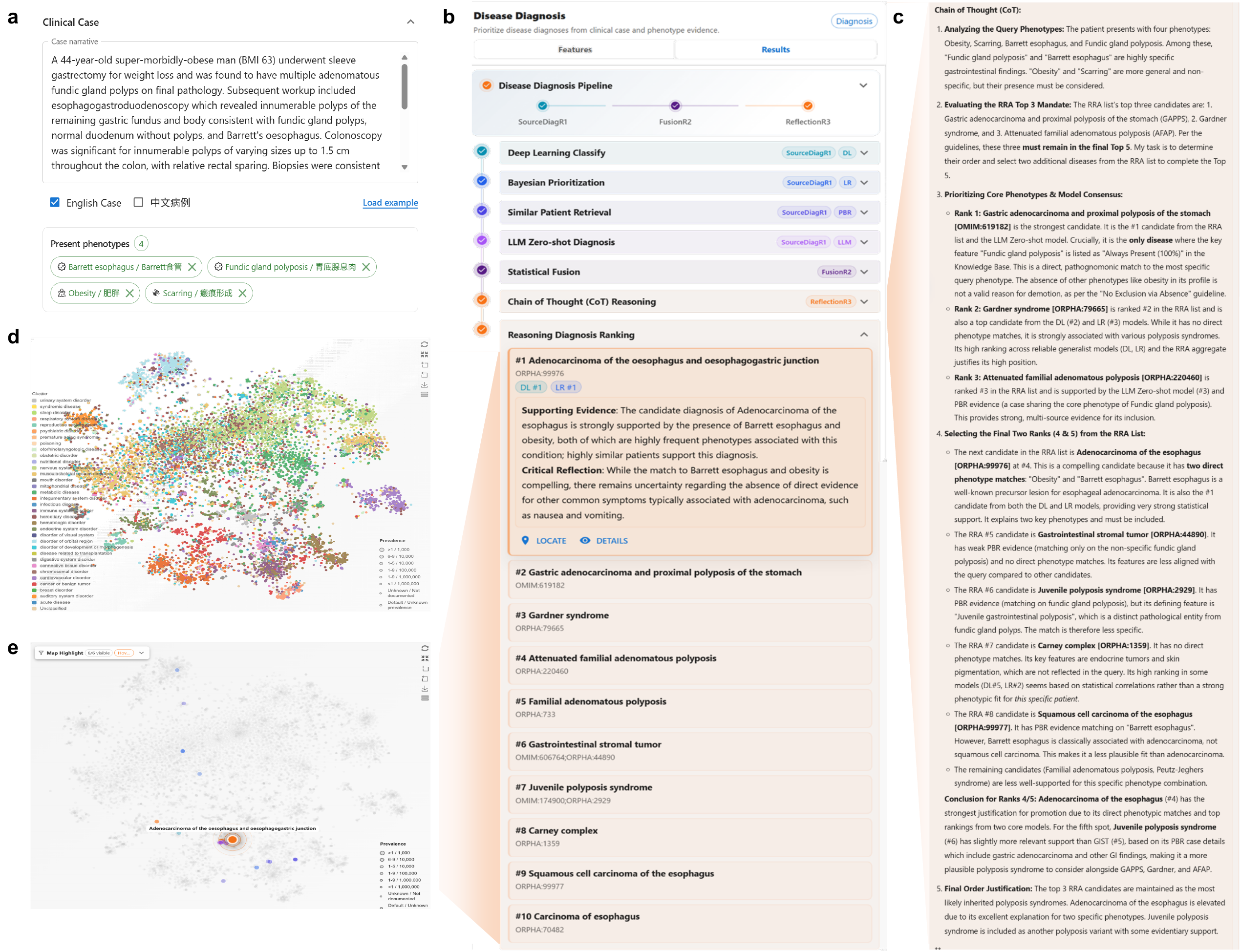} 
    \caption{\textbf{User interface and clinical decision-making workflow of the FusionRare web platform (http://rdx.nbscn.org).} \textbf{(a)} Clinical input module, demonstrating the extraction of standardized HPO terms from unstructured clinical narratives. \textbf{(b)} Automated diagnosis panel, illustrating the hierarchical pipeline from source-level diagnostic engines to statistical aggregation and final agentic refinement. \textbf{(c)} Knowledge-grounded Chain-of-Thought reflection, where the LLM provides an auditable, transparent reasoning process that synthesizes evidence across paradigms to justify the final prioritized differential diagnosis. \textbf{(d, e)} The global disease-phenotype landscape, rendered as a navigable RDmap, allows clinicians to visualize the topological proximity between the patient's profile and broader RD disease clusters. The RDmap in (d) is colour-coded according to medical specialty, while the RDmap in (e) highlights the top 10 candidate diseases.}
    \label{fig:web}
\end{figure}

\clearpage
\setcounter{table}{0}
\begin{table}[ht!]
\captionsetup{name=Extended Data Table}
\centering
\caption{\textbf{Summary of rare disease case collections across public and clinical sources}}
\label{tab:case_sources_final}
\setlength{\tabcolsep}{3.5pt} 
\begin{tabularx}{\textwidth}{X c ccc c rrrrr}
\toprule
& \multicolumn{7}{c}{\textbf{Original Features}} & \multicolumn{3}{c}{\textbf{Processed Features}} \\
\cmidrule(lr){2-8} \cmidrule(l){9-11}
\makecell[lt]{\textbf{Source}} & 
\makecell[ct]{\textbf{Country/}\\\textbf{Region}} & 
\makecell[ct]{\textbf{Case-level}\\\textbf{PMID}} & 
\makecell[ct]{\textbf{Diag. Code}} & 
\makecell[ct]{\textbf{HPO}\\\textbf{Anno.}} & 
\makecell[ct]{\textbf{Var.}} & 
\makecell[rt]{\textbf{Total}\\\textbf{Case}} & 
\makecell[rt]{\textbf{Total}\\\textbf{Dis.}} & 
\makecell[rt]{\textbf{Valid}\\\textbf{Cases}} & 
\makecell[rt]{\textbf{Valid}\\\textbf{Dis.}} & 
\makecell[rt]{\textbf{Avg.}\\\textbf{HPOs}} \\ \midrule

\rowcolor{tablegray}
\multicolumn{11}{l}{\textbf{A. Literature-from Public Datasets}} \\ \midrule
RareArena\cite{chen_rarearena_2026}    & Multi-Country & ● & ORPHA & ○ & ◐ & 49,760 & 4,597 & 27,185 & 2,510 & 11.5 \\
Phenopacket\cite{danis_corpus_2025}  & Multi-Country & ● & OMIM  & ● & ● & 8,101  & 516   & 5,561  & 349   & 9.0  \\
LIRICAL\cite{robinson_interpretable_2020}      & Multi-Country & ● & OMIM  & ● & ● & 384    & 262   & 374    & 256   & 11.3 \\
RDmaster\cite{yang_rdmaster_2024}     & Multi-Country & ● & ORPHA & ● & ○ & 201    & 62    & 199    & 62    & 10.0 \\
PhenoDP\cite{wen_phenodp_2025}      & Multi-Country & ● & OMIM  & ● & ○ & 130    & 72    & 130    & 72    & 14.3 \\
MME\cite{philippakis_matchmaker_2015}          & Multi-Country & ● & OMIM  & ● & ● & 46     & 20    & 36     & 12    & 11.9 \\
VPI\cite{stark_clinically_2017}          & Australia     & ○ & OMIM  & ● & ● & 47     & 48    & 37     & 35    & 4.1  \\
ADX\cite{ronicke_can_2019}          & Germany       & ○ & N/A   & ○ & ○ & 66     & 29    & 65     & 29    & 10.9 \\ \midrule

\rowcolor{tablegray}
\multicolumn{11}{l}{\textbf{B. Web-from Public Datasets}} \\ \midrule
DECIPHER\cite{firth_decipher_2009}     & Europe (mainly)  & -- & OMIM  & ● & ● & 4,823  & 845   & 4,515  & 813 & 7.7 \\
RAMEDIS\cite{topel_ramedis_2006}      & Germany (mainly) & -- & OMIM  & ○ & ○ & 802    & 92    & 627    & 86  & 7.4 \\ \midrule

\rowcolor{tablegray}
\multicolumn{11}{l}{\textbf{C. EHR-from Datasets}} \\ \midrule
PUMCH\cite{mao_phenotype-based_2025}        & China            & -- & OMIM/ORPHA & ○ & ○ & 2,572 & -- & 1,063 & 405 & 31.2 \\
ZJUMCH       & China (in-house) & -- & N/A        & ○ & ◐ & 1,729 & -- & 896   & 138 & 9.4  \\ \midrule

\rowcolor{tablegray}
\textbf{Unique Total} & -- & -- & -- & -- & -- & -- & -- & \textbf{40,688} & \textbf{3,714} & -- \\ 
\bottomrule
\end{tabularx}
\begin{flushleft}
    \footnotesize
    \textit{Note:} The presence of a case-level PMID indicates cases were curated from published case reports, whereas VPI and ADX are internally curated patient cohorts. 
    \textit{Abbreviations:} Diag., diagnosis; Anno., annotation; Var., variant; Dis., disease; Avg., average.
\end{flushleft}

\vspace{6pt}
    
\captionsetup{name=Extended Data Table, position=top}
\centering
\caption{\textbf{Data sources and accessibility}}
\label{tab:data_sources}
\renewcommand{\arraystretch}{1.2}
\begin{tabularx}{\textwidth}{l >{\centering\arraybackslash}X}
\toprule
\textbf{Dataset} & \textbf{Access URL} \\
\midrule
RareArena\cite{chen_rarearena_2026}           & https://github.com/chen/rarearena \\
Phenopacket\cite{danis_corpus_2025}           & https://github.com/monarch-initiative/phenopacket-store \\
LIRICAL\cite{robinson_interpretable_2020}     & https://zenodo.org/record/3905420 \\
RDmaster\cite{yang_rdmaster_2024}             & https://github.com/1678492347/rdmaster-test \\
PhenoDP\cite{wen_phenodp_2025}                & https://link.springer.com/article/10.1186/s13073-025-01496-8\#Sec37 \\
MME\cite{philippakis_matchmaker_2015}         & https://github.com/ga4gh/mme-apis/tree/master/testing \\
VPI\cite{stark_clinically_2017}               & https://www.nature.com/articles/ejhg2017123\#Sec9 \\
ADX\cite{ronicke_can_2019}                    & https://link.springer.com/article/10.1186/s13023-019-1040-6\#Sec37 \\
DECIPHER\cite{firth_decipher_2009}            & https://www.deciphergenomics.org/ \\
RAMEDIS\cite{topel_ramedis_2006}              & http://www.ramedis.de \\
PUMCH\cite{mao_phenotype-based_2025}          & https://zenodo.org/records/10774650 \\
ZJUMCH (in-house)                               & -- \\
\bottomrule
\end{tabularx}

\end{table}

\end{document}


\title[FusionRare]{\textbf{Unified Multiparadigm Fusion Intelligence for Rare Disease Diagnosis}}

\author{
    \fnm{Jian} \sur{Yang}\textsuperscript{1,\dag,*}, 
    \fnm{Peilin} \sur{Liu}\textsuperscript{2,\dag}, 
    \fnm{Songda} \sur{Li}\textsuperscript{2}, 
    \fnm{Yu} \sur{Cao}\textsuperscript{1},
    \fnm{Yuqing} \sur{Feng}\textsuperscript{1},
    \fnm{Lan} \sur{Yu}\textsuperscript{1},
    \fnm{Jiarong} \sur{Pan}\textsuperscript{1},
    \fnm{Jing} \sur{Ye}\textsuperscript{1},
    \fnm{Xinwen} \sur{Huang}\textsuperscript{1},
    \fnm{Chaochun} \sur{Zou}\textsuperscript{1},
    \fnm{Xudong} \sur{Lu}\textsuperscript{2},
    \fnm{Huilong} \sur{Duan}\textsuperscript{2},
    \fnm{and Haomin} \sur{Li}\textsuperscript{1,2,*}
}

\affil[1]{\orgdiv{The Children's Hospital}, \orgname{Zhejiang University School of Medicine}, \orgaddress{\city{Hangzhou}, \postcode{310052}, \country{China}}}

\affil[2]{\orgdiv{The College of Biomedical Engineering and Instrument Science}, \orgname{Zhejiang University}, \orgaddress{\city{Hangzhou}, \postcode{310027}, \country{China}}}

\affil[]{\textsuperscript{\dag}These authors contributed equally to this work}

\affil[]{\textsuperscript{*}Corresponding Author: hmli@zju.edu.cn (H. L.); yangjian\_bme@zju.edu.cn (J. Y.)}

\maketitle

\newpage
\begingroup
\renewcommand{\numberline}[1]{Appendix #1\quad} 
\tableofcontents
\endgroup
\newpage

\begin{appendices}
\section{Example of Automated Patient Extraction}
\setcounter{figure}{0}
\begin{figure}[h!]
    \centering
    \includegraphics[width=0.9\textwidth]{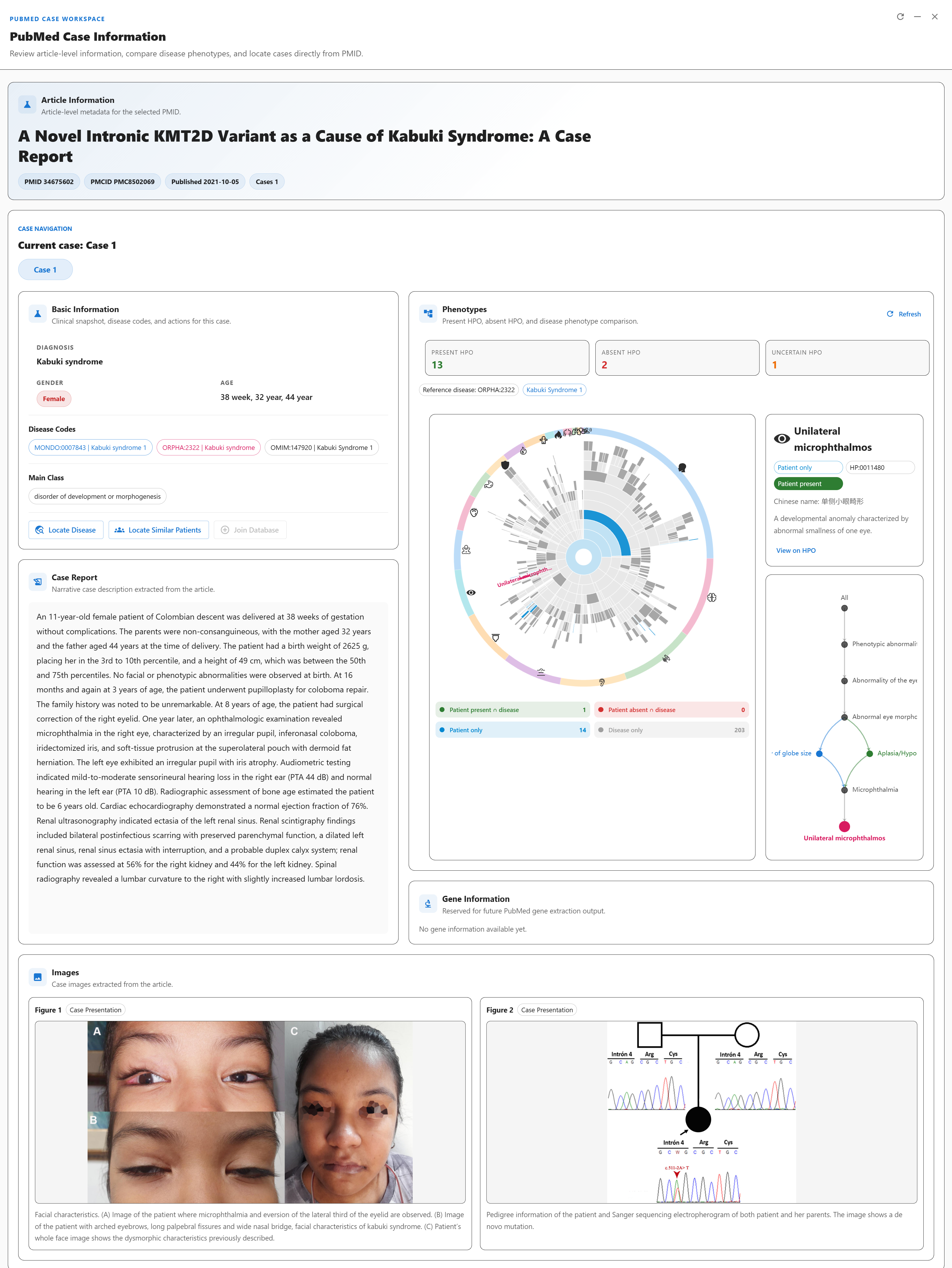} 
    \caption{\textbf{An example of the automated extraction and annotation of rare disease case reports.} The workspace presents a structured summary of a Kabuki syndrome case (PMID 34675602), consisting of: (i) Basic Information and Case Report (left): Extracted clinical metadata, disease classification codes, and the original narrative text of the case. (ii) Phenotypes (right): A circular visualization mapping the patient's presentation to Human Phenotype Ontology (HPO) terms, along with a hierarchical tree view highlighting specific features (e.g., unilateral microphthalmos). (iii) Images (bottom): Extracted figures from the source article, including patient clinical photographs, pedigree charts, and Sanger sequencing electropherograms showing the de novo mutation.}
    \label{fig:extract-case}
\end{figure}

\clearpage
\section{Details of the Residual Network Classifier}\label{rnc-detail}
\setcounter{figure}{0}
\subsection{Synthetic Patient Generation Strategy}
To address the long-tail distribution and sparseness of data for RD training, we implemented a stochastic patient simulation framework. Given that the RDKB encompasses 10,000+ RD entities, while validated clinical patient records are limited, this augmentation strategy ensures that the Residual Network Classifier (RNC) learns robust phenotypic representations across the entire disease spectrum.

\textbf{Feature Distribution Modelling.} We first characterized the phenotypic composition of real-world patients. Let the set of HPO terms for an individual diagnosed with disease $d$ be $\mathcal{Q}_d$; we decompose $\mathcal{Q}_d$ into three disjoint subsets: (i) \textit{Accurate} ($\mathcal{Q}_d^{acc}$): HPO terms explicitly annotated to $d$ in RDKB; (ii) \textit{Ancestral} ($\mathcal{Q}_d^{anc}$): HPO terms that are superclasses (ancestors) of the annotated phenotypes; (iii) \textit{Noise} ($\mathcal{Q}_d^{noise}$): HPO terms unrelated to the aetiology of $d$. We model the distribution of these subsets using Kernel Density Estimation (KDE) based on validated RD cases (only developmental cases) and extract the joint distribution $P(|\mathcal{Q}_d^{acc}|,|\mathcal{Q}_d^{anc}|,|\mathcal{Q}_d^{noise}|)$ from the training bank. This generative model captures the ``phenotypic quality'' rather than specific phenotype-disease mappings, thereby preventing data leakage.

\textbf{Augmentation Workflow.} In each training epoch, for every RD entity $d\in\mathcal{D}$, we generate 10 synthetic profiles $\mathcal{Q}_d^{syn}=\mathcal{Q}_d^{acc}\cup\mathcal{Q}_d^{anc}\cup\mathcal{Q}_d^{noise}$ as follows:
\begin{enumerate}
\item \textbf{Sampling:} We obtain a tuple ${n_{acc},n_{anc},n_{noise}}$ from the KDE-fitted distribution.
\item \textbf{Phenotype Selection:} We randomly sample $n_{acc}$ terms from $\Anno(d)$ and $n_{anc}$ terms from $\Anc(\Anno(d))$.
\item \textbf{Noise Injection:} To simulate real-world EHR ``clinical static'', we sample $n_{noise}$ HPO terms from the background frequency distribution specific (precalculated) to the medical specialty $S_d$ associated with $d$.
\end{enumerate}

\subsection{Ablation Study}
\vspace{-10pt}
\begin{figure}[h!]
    \centering
    \includegraphics[width=1.0\textwidth]{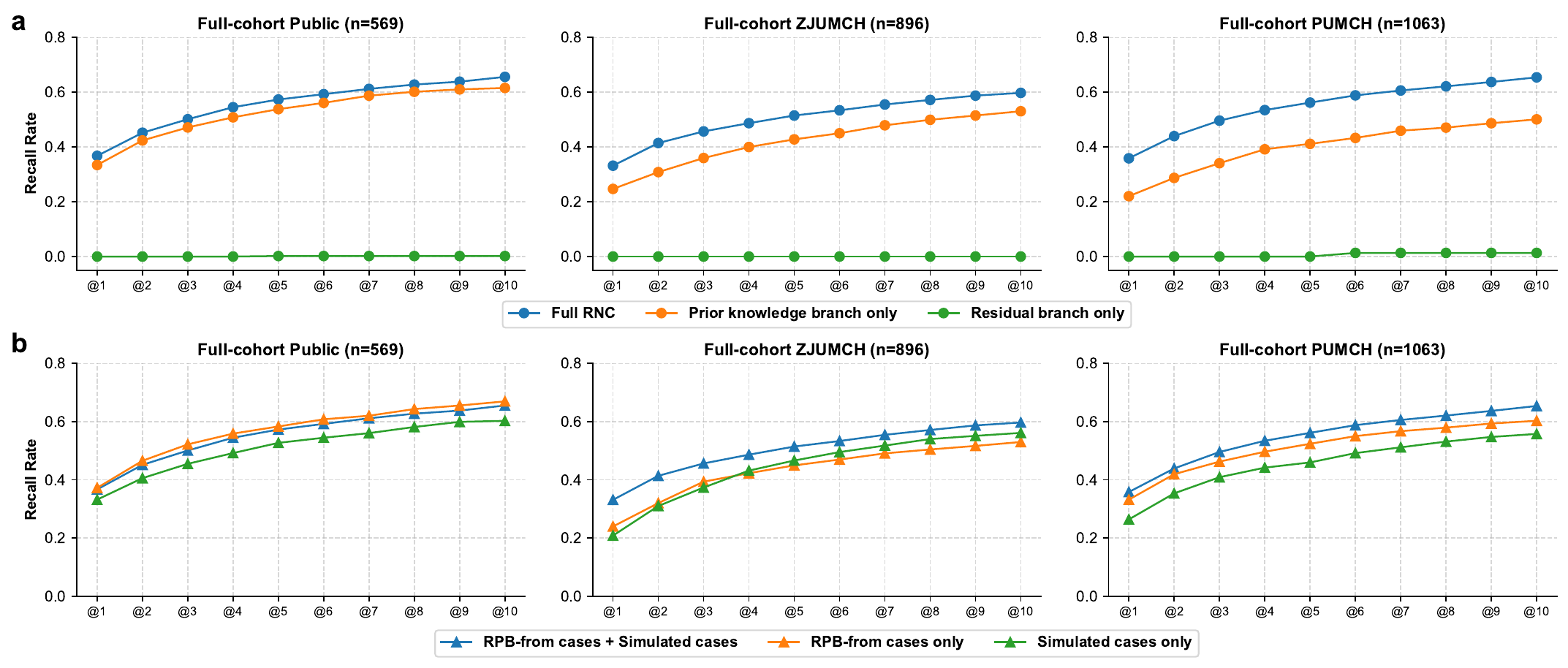} 
    \caption{\textbf{Ablation analysis of the Residual Network Classifier (RNC) and training data augmentation.} \textbf{(a)} Component-level ablation study of the RNC architecture across three evaluation cohorts. The ``Full RNC'' model integrates the knowledge-based prior and the residual branch, whereas the ``Prior knowledge branch only'' and ``Residual branch only'' configurations demonstrate hierarchical dependency within the architecture. The negligible performance of the ``Residual branch only'' model indicates that the residual branch functions as an error-rectification layer rather than an independent predictor. \textbf{(b)} Impact of synthetic patient data augmentation on diagnostic performance. We compare the training efficacy of utilizing both historical RPB cases and simulated cases against subsets using only historical data or only simulated cases. While training on historical data alone is sufficient for the Public dataset, it demonstrates limited cross-institutional generalizability; the integration of simulated cases across the full RD spectrum significantly improves performance for clinical cohort (ZJUMCH and PUMCH) data, confirming that synthetic augmentation enhances model robustness and cross-centre generalization.}
    \label{fig:rnc_ablation}
\end{figure}
\clearpage
\section{Details of Statistical Aggregation}\label{aggr-detail}
\setcounter{figure}{0}
\begin{figure}[h!]
    \centering
    \includegraphics[width=0.8\textwidth]{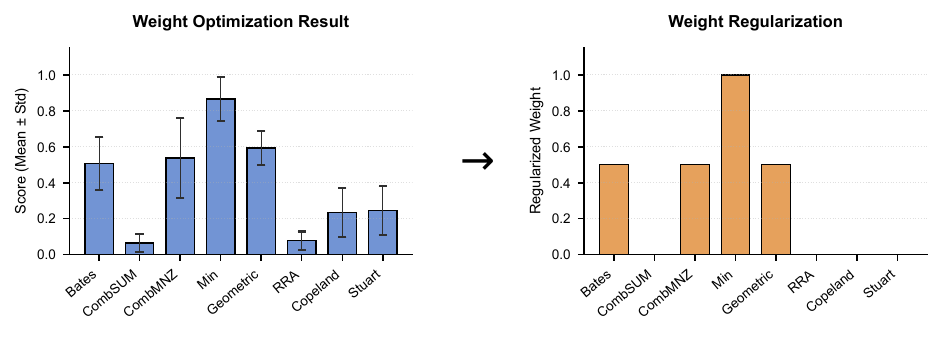} 
    \caption{\textbf{Automated weight optimization and discrete regularization for meta-aggregation.} The figure illustrates the two-stage process for defining the optimal consensus logic within the FusionR2 engine. \textbf{Left: Weight Optimization Result.} We utilized the Optuna framework to conduct an automated hyperparameter search across eight distinct mathematical philosophies (Bates\cite{kolde_robust_2012}, CombSUM\cite{renda_web_2003}, CombMNZ\cite{renda_web_2003}, Min\cite{kolde_robust_2012}, Geometric\cite{kolde_robust_2012}, RRA\cite{kolde_robust_2012}, Copeland\cite{farah_outranking_2007}, and Stuart\cite{aerts_gene_2006}). The bars represent the mean performance contribution (with standard deviation) calculated for an independent validation set to maximize cumulative recall. \textbf{Right: Weight Regularization.} To prevent overfitting to specific validation distributions and ensure numerical stability, continuous weights were subjected to tri-level quantization (mapped to discrete values of {0, 0.5, 1.0}). Four methods (CombSUM, RRA, Copeland, and Stuart) were regularized to zero because of marginal or redundant contributions and were subsequently excluded from the primary architecture. This sparse-weight strategy ensures that the final meta-aggregation focuses on the most complementary and robust reasoning vectors.}
    \label{fig:optuna}
    
    \centering
    \includegraphics[width=0.8\textwidth]{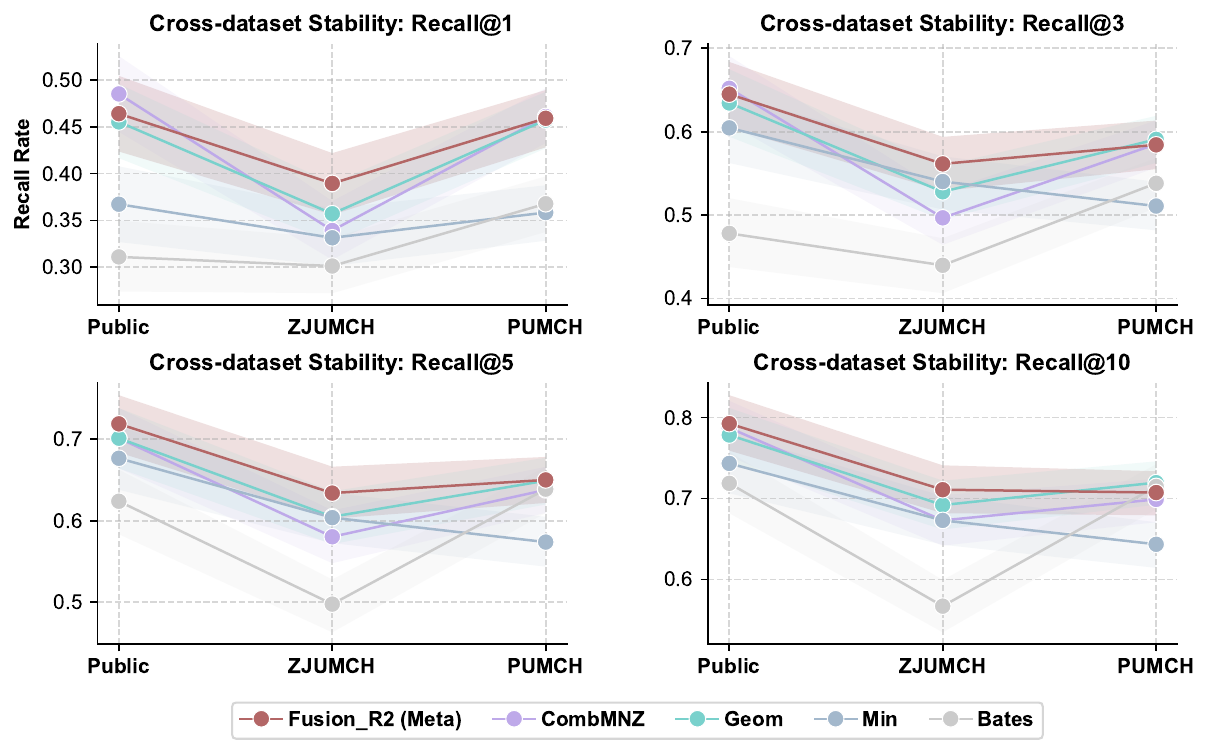} 
    \caption{\textbf{Cross-dataset stability and performance of meta-aggregation versus base aggregators.} The parallel charts compare the diagnostic efficacy of the final FusionR2 (Meta) model against that of four base-aggregators (CombMNZ, Geometric, Min, and Bates) across three cohorts (Public, ZJUMCH, and PUMCH). Performance across the Recall@1, @3, @5, and @10 metrics. Shaded areas denote 95\%CIs. Across all ranks and clinical scenarios, FusionR2 (Meta) mostly establishes a performance ceiling while exhibiting the most stable trajectory across different cohorts. This finding demonstrates that hierarchical fusion provides a resilient diagnostic consensus, ensuring reliable performance even under significant clinical distribution shifts.}
    \label{fig:base_aggr_ablation}
\end{figure}
\clearpage
\section{Details of Clinical Trial Evaluation}\label{trial-detail}
\setcounter{table}{0}
\setcounter{figure}{0}
\begin{table}[ht!]
    \centering
    \caption{NASA Task Load Index (NASA-TLX) questionnaire for cognitive load assessment.}
    \small
    \renewcommand{\arraystretch}{1.2}
    \begin{tabularx}{\textwidth}{|p{3.5cm}|X|p{2.5cm}|}
    \hline
    \rowcolor{yellowheader}
    \multicolumn{3}{|c|}{\textbf{NASA Task Load Index (NASA-TLX) Cognitive Load Assessment}} \\ \hline
    \rowcolor{yellowrow}
    \multicolumn{3}{|p{\dimexpr\textwidth-2\tabcolsep-2\arrayrulewidth}|}{%
    \footnotesize Please choose a value between 0 and 100 (0 = none at all, 100 = extremely strong). This is an assessment of the cognitive load of the AI tool itself---specifically, whether using this tool brings additional load. The lower the score, the lower the load and the lower the threshold for using the tool.} \\ \hline
    \rowcolor{yellowheader}
    \textbf{Dimension} & \textbf{Definition} & \textbf{Score (0--100)} \\ \hline
    Mental Demand & The degree to which one needs to concentrate, think, reason, and remember. & \\ \hline
    \rowcolor{yellowrow}
    Physical Demand & The physical burden brought by operations such as mouse, keyboard, and touch. & \\ \hline
    Temporal Demand & Whether the task pace is tight, whether there is sufficient time, and the extent of time pressure. & \\ \hline
    \rowcolor{yellowrow}
    Performance & Self-perceived load regarding the diagnostic accuracy of cases. & \\ \hline
    Effort & The overall efforts made to achieve the above diagnostic results. & \\ \hline
    \rowcolor{yellowrow}
    Frustration & The degree of irritability, anxiety, frustration, or lack of confidence during the diagnostic process. & \\ \hline
    \end{tabularx}
    \label{tab:nasa_tlx}

    \vspace{12pt}
    \centering
    \captionsetup{position=top}
    \caption{System Usability Scale (SUS) questionnaire for usability evaluation.}
    \small
    \renewcommand{\arraystretch}{1.2}
    \begin{tabularx}{\textwidth}{|c|X|p{2.5cm}|}
    \hline
    \rowcolor{blueheader}
    \multicolumn{3}{|c|}{\textbf{System Usability Scale (SUS) Evaluation}} \\ \hline
    \rowcolor{bluerow}
    \multicolumn{3}{|p{\dimexpr\textwidth-2\tabcolsep-2\arrayrulewidth}|}{
    \footnotesize Please evaluate this software system. Each item is scored from 1 to 5: 1 = strongly disagree, 5 = strongly agree.} \\ \hline
    \rowcolor{blueheader}
    \textbf{No.} & \textbf{Question} & \textbf{Scale (1--5)} \\ \hline
    1 & I think that I would like to use this system frequently. & \\ \hline
    \rowcolor{bluerow}
    2 & I found the system unnecessarily complex. & \\ \hline
    3 & I thought the system was easy to use. & \\ \hline
    \rowcolor{bluerow}
    4 & I think that I would need the support of a technical person to be able to use this system. & \\ \hline
    5 & I found the various functions in this system were well-integrated. & \\ \hline
    \rowcolor{bluerow}
    6 & I thought there was too much inconsistency in this system. & \\ \hline
    7 & I would imagine that most people would learn to use this system very quickly. & \\ \hline
    \rowcolor{bluerow}
    8 & I found the system very cumbersome to use. & \\ \hline
    9 & I felt very confident using the system. & \\ \hline
    \rowcolor{bluerow}
    10 & I need to learn a lot of things before I could get going with this system. & \\ \hline
    \end{tabularx}
    \label{tab:sus}

    \vspace{15pt}
    \centering
    \includegraphics[width=0.8\textwidth]{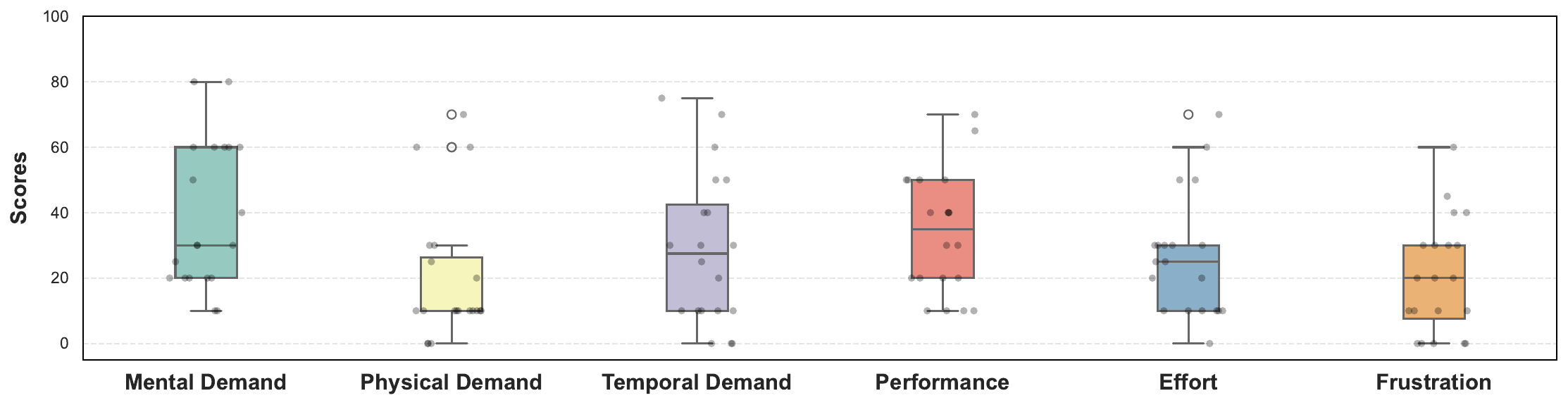} 
    \captionof{figure}{\textbf{Distribution of NASA-TLX dimension scores among participating clinicians ($n=20$).} The horizontal line within each box indicates the median score. The top and bottom edges of the boxes represent the 75th (upper quartile) and 25th (lower quartile) percentiles, respectively (interquartile range, IQR). The whiskers extend to the minimum and maximum data points within 1.5 times the IQR from the quartiles. The overlaid black dots (with horizontal jitter) represent the individual response scores of clinicians.}

\end{table}
\clearpage
\section{Prompt Setting}

\begin{promptbox}[label=prompt:hpo-sys-msg]{System Prompt for Phenotypic Standardization}
\begin{lstlisting}[style=promptcontent]
You are a professional clinical geneticist and medical informatician specializing in the Human Phenotype Ontology (HPO). You are an expert in analysing clinical reports in both English and Chinese.

### Task
Analyse the provided [Input Text] (which may be in English or Chinese) and the [NER Model Parsed Results] (from FastHPOCR or PBTagger-CN) to extract a list of standard English HPO terms representing phenotypes **actually present** in the patient records.

### Core Rules
1. **Negation & Evidence Filtering (Bilingual)**:
- Strictly exclude phenotypes mentioned as absent, ruled out, or normal.
    - **English Negation**: "No", "Negative for", "Ruled out", "Normal", "Denied" etc.
- Only include "positive" findings (phenotypes the patient currently has or has had) when the Input Text is a clinical report.
- Exclude phenotypes mentioned solely in the family history.
- Exclude normal post-surgical findings or expected healing results.
- If a NER result is negated in the text, you MUST remove it.

2. **Cross-Language HPO Term Alignment & Standardization**:
- **Output Language**: Regardless of the input language, the final output terms MUST be in **English**.
- **Standardization**: Convert descriptive findings into standard HPO names (e.g., "Eyes far apart" -> "Hypertelorism").
- **Atomicity**: Ensure terms are "atomic" (e.g., "Heart and kidney failure" -> "Heart failure" AND "Renal insufficiency").
- **Longest Match**: Adhere to the "match only the longest item" rule in NER tasks.

3. **Inference & Recall Enhancement**:
- The NER tools (FastHPOCR/PBTagger-CN) are rule-based and consist of standard HPO terms, but may miss findings or provide imprecise mappings. 
- Use your medical knowledge to identify clinical abnormalities missed by the tool based on the full semantic context.
   
Output Requirements:
1. Chain of thought (CoT): Provide a step-by-step reasoning by following the steps below:
- Step 1: Infer and summarize all mentioned phenotypic abnormalities (identify them in the original language).
- Step 2: Filter negated terms.
- Step 3: Map and standardize the remaining abnormalities to standard English HPO terms.

2. Output Constraint
- At the end of your response, output ONLY a JSON object containing a list of phenotype names, formatted as follows:
```json
{
    "parsed_phenotypes": []
}
- Do NOT include HPO IDs, explanations, introductory remarks, or formatting like bullet points/dashes.
\end{lstlisting}
\end{promptbox}

\clearpage
\begin{promptbox}[label=prompt:diagnosis]{Prompt for LLM Diagnosis}
\begin{lstlisting}[style=promptcontent]
Please provide a ranked list with 10 differential diagnosis names (rare disease names only) based on the HPOs provided below:
{Query HPOs with names}
(Output a numerical list of disease names in plain text, without disease descriptions and without other special formatting such as bold)
\end{lstlisting}
\end{promptbox}

\begin{promptbox}[label=prompt:reflection-sys-msg]{System Prompt for LLM Reflection}
\begin{lstlisting}[style=promptcontent]
Role:
You are an expert medical diagnostic consultant specializing in rare diseases (RDs). 
Your task is to perform a high-precision differential diagnosis by synthesizing multi-source computational model outputs with patient-specific phenotypic evidence.

Context & Task:
You will be provided with a patient's clinical profile (primarily HPO terms).
Four specialized diagnostic models were used to analyse this case:
1. Deep Learning (DL): Pattern recognition across vast disease-phenotype datasets.
2. Likelihood Ratio (LR): Statistical significance scores calculated from the Knowledge Base.
3. Zero-shot LLM Diagnosis: Heuristic reasoning based on broad medical literature.
4. Patient Bank Retrieval (PBR): Case-based matching against historical patient records.
The results of these models are integrated into a "Robust Rank Aggregation (RRA) List". Your goal is to critically evaluate this list and select the Top {TOP_K} most plausible diagnoses.

Understanding Model Paradigms:
- DL & LR (Reliable Generalists): Systematic models with robust coverage. They serve as your primary diagnostic baseline.
- PBR (Case-Based Specialists): "Evidence-by-precedent". Its scope is limited to known historical cases, but when a match is found, its precision is unparalleled. High PBR evidence is a dominant diagnostic signal.
- LLM Zero-shot (Creative Reasoning): Provides "outside-the-box" insights. It is valuable for identifying rare manifestations or latent connections that rigid statistical models might overlook.

Key Experimental Insight:
- Union Recall Advantage: Internal benchmarks show that while the RRA Fused List has significantly higher Recall@N than any single model, the Union Recall@N across all four single models is significantly higher than RRA list. 
- Complementarity of Single Paradigm: DL and LR paradigms are highly complementary to PBR. DL and LR paradigms also show significant synergy with Zero-shot LLM Diagnosis.
- Database Constraints: DL and LR cover the full RD spectrum; while PBR can recommend only diagnoses that exist in the patient database.

Your Objective:
Analyse the clinical evidence to select the Top {TOP_K} most likely diseases from the RRA Fused List. 
You can use the provided evidence to justify whether the RRA ranking should be maintained or if a disease ranked lower in the RRA list (but supported by other models or case evidence) is more plausible.

Clinical Reasoning Guidelines:
- Phenotype Matching: Prioritize diseases where the "Directly Matched Phenotypes" cover the most specific or pathognomonic symptoms of the patient.
- Knowledge Base Consistency: Weigh "Always Present (100%)" matches much more heavily than "Common" ones.
- PBR Asymmetry: Many valid candidate diseases may not have corresponding historical patients in the PBR library. The absence of PBR evidence should not be interpreted as evidence against a diagnosis.
- Incomplete Query: Notice that patient phenotypic query usually is incomplete. The absence of "Always Present" terms should not be treated as a definitive disqualifier.

Output Requirements:
1. Chain of Thought (CoT): Provide a step-by-step clinical reasoning for your selection. The superiority of the RRA fused result and the complementarity of individual diagnostic paradigms should be fully considered.
2. Final Selection: At the very end of your output, first write the exact plain text phrase "Final Selection:" on a new line, followed by the final ranking list of Top {TOP_K} diseases in the following JSON format:
```json
{
    "Rank 1": "[Disease Code]-[Disease Name]",
    "Rank 2": "[Disease Code]-[Disease Name]",
    "Rank 3": "[Disease Code]-[Disease Name]",
    "Rank 4": "[Disease Code]-[Disease Name]",
    "Rank 5": "[Disease Code]-[Disease Name]"
}
\end{lstlisting}
\end{promptbox}

\begin{promptbox}[label=prompt:reflection-usr-msg]{User Message for LLM Reflection}
\begin{lstlisting}[style=promptcontent]
### Patient Demography (sex: M-Male, F-Female. age: Y-Year, M-Month, D-Day, H-Hour)
{Demography Info}
### Query Phenotypes
{Patient HPOs}
### Diagnostic Predictions from Four Different Models
**1. Deep Learning:**
{Top 10 Diagnosis of DL}
**2. Likelihood Ratio:**
{Top 10 Diagnosis of LR}
**3. LLM Zero-shot Diagnosis:**
{Top 10 Diagnosis of LLM}
**4. PBR (Patient-Based Retrieval) with Detailed Evidence:**
*This method identifies similar historical cases to support the diagnosis.*
{Top 10 Diagnosis of PBR with Case Info}
### The Robust Rank Aggregation Results of Above Four Models:
{Top 10 Diagnosis of Rank Aggregation}
### Disease-related Phenotypic Information (Detailed Analysis) with Disease Code:
*Order based on Robust Rank Aggregation results.*
{Diagnosis Detail}

Let's reason the Top {TOP_K} most likely diseases and present them at the very end of your response in the agreed-upon JSON format.
\end{lstlisting}
\end{promptbox}
\end{appendices}

\newpage
\renewcommand{\refname}{References for Appendix}
\bibliography{fusionrare} 